\documentclass[11pt]{article}

\usepackage[preprint]{acl}

\usepackage{times}
\usepackage{latexsym}

\usepackage[T1]{fontenc}

\usepackage[utf8]{inputenc}

\usepackage{microtype}

\usepackage{inconsolata}

\usepackage{graphicx}

\usepackage{dirtytalk}
\usepackage{enumitem}
\usepackage{url}
\usepackage{amsmath}
\usepackage{booktabs}
\usepackage{float}

\usepackage{multirow}

\usepackage[ruled,vlined,linesnumbered,norelsize]{algorithm2e}
\SetKwInOut{Input}{Input}
\SetKwInOut{Output}{Output}

\usepackage{tikz}
\usepackage{circuitikz}
\usetikzlibrary{positioning, shapes.geometric, arrows}
\tikzstyle{container} = [rectangle, draw=black, fill=gray!20, text centered, rounded corners, minimum height=1cm]
\tikzstyle{process} = [rectangle, draw=blue, fill=blue!20, text centered, rounded corners, minimum height=1cm]
\tikzstyle{data} = [rectangle, draw=yellow!50!black, fill=yellow!50, text centered, rounded corners, minimum height=1cm]
\tikzstyle{storage} = [rectangle, draw=teal, fill=teal!20, text centered, rounded corners, dashed, minimum height=1cm]
\tikzstyle{arrow} = [thick, ->, >=stealth]
\tikzstyle{dashed_arrow} = [thick, dashed, ->, >=stealth]
\tikzstyle{note} = [rectangle, draw=black, fill=white, text centered, rounded corners, minimum height=1cm]

\title{A Systematic Exploration of Text Decomposition and Budget Distribution in Differentially Private Text Obfuscation}

\author{Stephen Meisenbacher, Angelo Kleinert, \and Florian Matthes \\
Technical University of Munich\\
School of Computation, Information and Technology \\
Department of Computer Science\\
Garching, Germany\\
\texttt{\{stephen.meisenbacher,angelo.kleinert,matthes\}@tum.de} \\
}

\begin{document}
\maketitle
\begin{abstract}
The goal of \textit{differentially private text obfuscation} is to obfuscate, or \say{perturb}, input texts with Differential Privacy (DP) guarantees, such that the private output texts are quantifiably indistinguishable from the originals. While perturbation at the word level is intuitive, meaningful text privatization happens on complete documents. Recent research has laid the groundwork for reasoning about \textit{privacy budget distribution}, namely, how an overall $\varepsilon$ budget can be sensibly distributed among the component pieces of a text. We perform a systematic evaluation of multiple text decomposition and budget distribution techniques in the context of DP text obfuscation, testing how different methods for chunking texts can be combined with techniques for allocating $\varepsilon$ to these chunks. Our experiments reveal that such design choices are very important, as even with comparable privacy budgets, significantly different results can occur based on which methods are chosen. In this, we provide credible evidence of the feasibility of maximizing empirical trade-offs by optimizing DP obfuscation procedures.
\end{abstract}

\section{Introduction}
In the rapidly advancing world of AI and LLMs, the reliance of the modern technological economy on massive data collection has justifiably raised concerns of privacy \cite{yao2024survey}. Calls for privacy protection have been met by a significant string of privacy research in the context of Natural Language Processing (NLP) \cite{9152761,9592788,sousa2023keep}. One theoretically viable, but practically challenging solution comes in the form of text privatization under Differential Privacy (DP) guarantees \cite{igamberdiev-etal-2022-dp}, thus spring-boarding numerous works at the intersection of DP and NLP \cite{hu-etal-2024-differentially}.

One of the immediate challenges of privatizing text under DP comes with reasoning about the \textit{unit of privatization} \cite{klymenko-etal-2022-differential}. Many of the solutions to this challenge propose DP text obfuscation at the sub-document level, such as with the perturbation of words \cite{feyisetan_balle_2020} or tokens \cite{mattern-etal-2022-limits}, thereby creating a new challenge of meaningful text privatization for downstream usage. Relying on the compositionality of DP \cite{dwork2006differential} -- whereby repeated perturbations on the same (text) data are \textit{composed} \cite{feyisetan_balle_2020} -- recent works have begun to conduct evaluations of sub-document-level perturbation mechanisms on full documents \cite{meisenbacher-etal-2024-collocation}. This is achieved by assigning a privacy budget (the $\varepsilon$ parameter of DP) to the document, and subsequently allocating this budget among component pieces (e.g., words).

\begin{figure*}[ht!]
    \centering
    \includegraphics[width=0.97\linewidth]{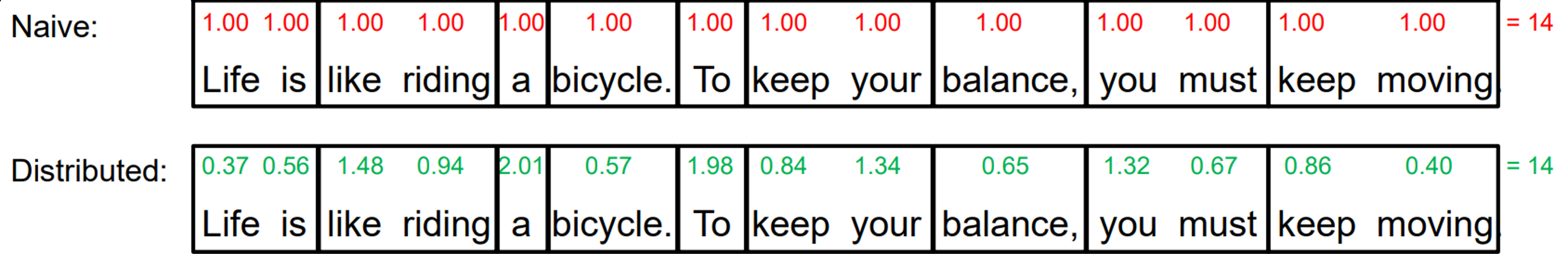}
    \caption{An example of text decomposition and budget distribution for DP text obfuscation. Given the same input text of 14 words (excluding punctuation), one can \textit{decompose} the text into meaningful chunks beyond simple word tokenization. In addition, with a privacy budget of 14, one can either ``naively'' (uniformly) allocate it, or rather, distribute the budget sensibly such that more important words receive higher privatization (lower $\varepsilon$), and vice versa. Note: for understandability, we illustrate word-level budgets; we simply add these for the chunk-level budget.}
    \label{fig:example}
\end{figure*}

While a simple approach may be to evenly distribute a privacy budget to all component words of a document (top of Figure \ref{fig:example}), recent work has proposed techniques for more intelligently performing this allocation (bottom of Figure \ref{fig:example}) \cite{10.1145/3714393.3726504}. Similarly, other work demonstrates the value in more complex \textit{text decomposition} methods, where an input document is \textit{chunked} dynamically, for example by phrases and n-grams \cite{NEURIPS2021_28ce9bc9,meisenbacher-etal-2024-collocation}, or even sentences \cite{meehan-etal-2022-sentence}. The combination of these techniques, i.e., dynamic text decomposition \textit{and} privacy budget distribution, has neither been explored nor systematically evaluated.

We design a systematic evaluation of text decomposition and privacy budget distribution with DP text obfuscation, asking the following questions:

\begin{enumerate}[leftmargin=1.1cm]
    \itemsep -0.3em
    \item[RQ1.] How can methods for text decomposition and privacy budget allocation be combined and evaluated for DP text obfuscation?
    \item[RQ2.] Are there significant differences, particularly in the privacy-utility trade-offs, between the combinations of RQ1?
\end{enumerate}

Designing an experimental setup with two datasets, three privatization levels ($\varepsilon$), five decomposition methods and six distribution methods, we comparatively analyze \textbf{180} different DP text obfuscation configurations, evaluating these setups on privacy, utility, and the trade-offs between the two.

While we find that there does not exist a universally dominant combination of methods in terms of privacy \textit{and} utility, our results demonstrate that privacy, utility, \textit{or} trade-offs can be optimized depending on which decomposition-distribution setup is chosen. These findings are supported by significant differences in metrics between setups, showcasing that such design choices \textit{do} matter and have quantifiable and practically relevant implications.

Our work advances the field of DP text obfuscation by comprehensively exploring different avenues for transforming local DP perturbations into usable and optimized privatized documents. In particular, we make the following contributions:

\begin{enumerate}
    \itemsep -0.3em
    \item We systematically evaluate 180 setups for DP text obfuscation, under various text decomposition and budget distribution schemes.
    \item We perform a comparative analysis of privacy and utility preservation, demonstrating significant differences between setups.
    \item We open-source our modular code at {\small\url{https://github.com/sjmeis/DP-Decompose-Distribute}}.
\end{enumerate}

\section{Foundations and Related Work}
\paragraph{DP and Text Privatization.}
DP \cite{dwork2006differential} ensures a level of formal privacy by bounding the contribution of any individual in a dataset to queries or computations performed on the data.
This is governed by the privacy parameter ($\varepsilon$, or the \textit{privacy budget}), which affords \textit{indistinguishability} to the individual. Formally, this is expressed as:
\vspace{-1pt}
\[
    Pr[\mathcal{M}(D) \in \mathcal{S}]
    \leq e^{\varepsilon} Pr[\mathcal{M}(D') \in \mathcal{S}],
\]

\noindent for any databases $D$ and $D'$ differing in exactly one element (or \say{individual}), any $\varepsilon > 0$, any function $\mathcal{M}$, and all $\mathcal{S} \subseteq Range(\mathcal{M})$.

In the context of text privatization, where DP is applied directly on the data itself, the notion of \textit{local} DP \cite{4690986} is often adopted. In this paradigm, the task of achieving DP is shifted to the user level, instead of some central curator. This, however, imposes a stricter indistinguishability requirement, as applying local DP considers the entire \textit{universe} of data values rather than those contained within the dataset $D$ of central (global) DP. As such, the DP notion becomes:
\vspace{-2pt}
\[
Pr[\mathcal{M}(x) = z] \le e^\varepsilon Pr[\mathcal{M}(x') = z]
\]
         
\noindent Thus, an observed output cannot be attributed to \textit{any} data point within a bounded probability.

The local DP paradigm is useful for applications in text privatization, as it allows for local perturbations of text representations, where individual data points can be likened to units of language, such as words or documents \cite{klymenko-etal-2022-differential}.

\paragraph{Metric Local DP.}
Some of the first proposed solutions to bring DP to the task of text privatization leveraged a generalized notion called \textit{metric} (local) DP (MLDP) \cite{chatzikokolakis2013broadening}. MLDP relaxes the strict requirement of local DP, namely the requirement of indistinguishability between \textit{any} two data points, by scaling the privacy loss based on proximity in some metric space (formally, by scaling $\varepsilon$ by some distance metric $d$). Early works in DP text privatization recognized the adaptability of MLDP to embedding spaces \cite{fernandes2019generalised,feyisetan_balle_2020}, e.g., by \textit{obfuscating} words via their embedding representations.

Applications of MLDP to text take one of two forms: direct obfuscation of embedding representations \cite{feyisetan_balle_2020} or modeling text replacement as a selection problem \cite{yue-etal-2021-differential}. We focus on the former, which views obfuscation holistically and does not constrain replacement candidates to a finite set. In particular, MLDP perturbations typically operate in three steps \cite{10.1145/3746252.3760888}: (1) \textit{embedding}, (2) \textit{perturbation} via DP noise, and (3) \textit{projection} to a replacement (word). While most DP text obfuscation mechanisms follow this general paradigm, MLDP allows for flexibility in the mechanism's design, resulting in numerous works that improve the usability of DP outputs \cite{carvalho2023tem,arnold-etal-2023-driving,arnold-etal-2023-guiding,10.1145/3746252.3760888}. Extending beyond word-level perturbations, \citet{feyisetan_balle_2020} provide a framework for \textit{document-level} obfuscation via basic DP composition.

Particularly in the evaluation of MLDP text obfuscation approaches, \citet{10.1145/3714393.3726504} highlight an important consideration with word-level approaches, namely the need to test with uniform document-level privacy budgets; otherwise, comparability is lacking if these budgets are not equal (i.e., when composition of word-level budgets is unbounded). We ground our work in this, as we seek to optimize the allocation of a fixed privacy budget among the components of a text.

\paragraph{Text Chunking and Multi-Word Expressions.}
The task of text chunking is an early NLP problem that aims to segment texts into coherent chunks \cite{beeferman1999statistical}. While word segmentation is the simplest approach, other techniques logically separate phrases or sentences \cite{pak2017text}. \citet{pecina-schlesinger-2006-combining} survey 82 association measures for extracting \textit{collocations}, or meaningful groupings of words that often appear together. More generally, multi-word expressions (MWEs) \cite{sag2002multiword} have been widely studied \cite{constant-etal-2017-survey}, pointing to the importance of linguistic units beyond word boundaries.

Our work draws motivation from DP text privatization research operating between the word and document level, such as with collocations \cite{meisenbacher-etal-2024-collocation} or sentences \cite{meehan-etal-2022-sentence}. We focus on MWEs, as they strike a balance between words, which lack context, and sentences, which often are not cohesive units of expression.

\section{Methodology}
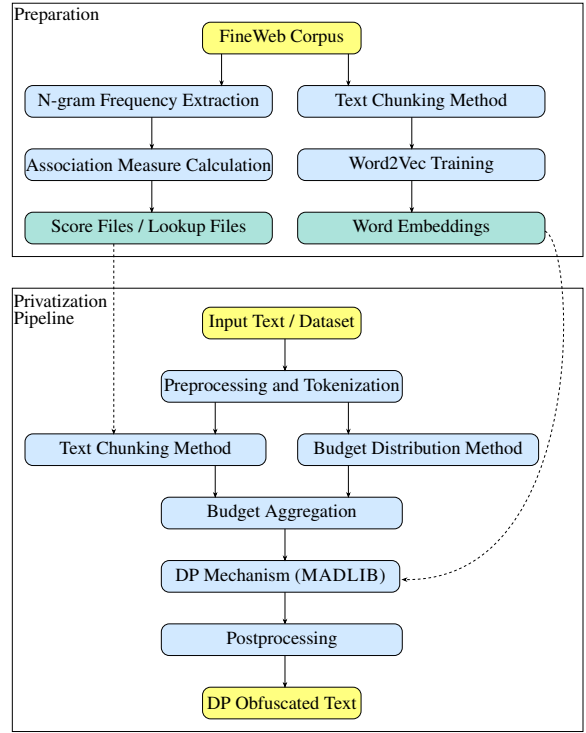
\begin{figure}[t!]
\centering
\resizebox{0.48\textwidth}{!}{%
\begin{circuitikz}
\tikzstyle{every node}=[font=\huge]
\draw  (0,4) rectangle (22.5,-13.5);
\draw  (0,15.25) rectangle (22.5,5.25);
\node [font=\huge] at (1.9,3.5) {Privatization};
\draw [ fill={rgb,255:red,255; green,255; blue,128} , rounded corners = 9.0] (7.5,14.5) rectangle  node {\huge FineWeb Corpus} (13.75,13.25);
\draw  (8.25,13.5) rectangle (8.25,13.5);
\draw  (6,14) rectangle (6,14);
\draw  (10.25,14.5) rectangle (10.25,14.5);
\draw  (10.25,14.5) rectangle (10.25,14.5);
\draw  (6.25,11.25) rectangle (6.25,11.25);
\draw [ fill={rgb,255:red,210; green,233; blue,255} , rounded corners = 9.0] (0.5,12) rectangle  node {\huge N-gram Frequency Extraction} (10.25,10.75);
\draw [ fill={rgb,255:red,210; green,233; blue,255} , rounded corners = 9.0] (11.25,12) rectangle  node {\huge Text Chunking Method} (21,10.75);
\draw [ fill={rgb,255:red,210; green,233; blue,255} , rounded corners = 9.0] (11.25,9.5) rectangle  node {\huge Word2Vec Training} (21,8.25);
\draw [ fill={rgb,255:red,210; green,233; blue,255} , rounded corners = 9.0] (0.5,9.5) rectangle  node {\huge Association Measure Calculation} (10.25,8.25);
\draw [ fill={rgb,255:red,172; green,227; blue,219} , rounded corners = 9.0] (0.5,7) rectangle  node {\huge Score Files / Lookup Files} (10.25,5.75);
\draw [ fill={rgb,255:red,172; green,227; blue,219} , rounded corners = 9.0] (11.25,7) rectangle  node {\huge Word Embeddings} (21,5.75);
\draw [ fill={rgb,255:red,255; green,255; blue,128} , rounded corners = 9.0] (7.5,3.25) rectangle  node {\huge Input Text / Dataset} (13.75,2);
\draw [ fill={rgb,255:red,210; green,233; blue,255} , rounded corners = 9.0] (5.875,0.75) rectangle  node {\huge Preprocessing and Tokenization} (15.375,-0.5);
\draw [ fill={rgb,255:red,210; green,233; blue,255} , rounded corners = 9.0] (0.5,-1.75) rectangle  node {\huge Text Chunking Method} (10,-3);
\draw [ fill={rgb,255:red,210; green,233; blue,255} , rounded corners = 9.0] (11.25,-1.75) rectangle  node {\huge Budget Distribution Method} (20.75,-3);
\draw [ fill={rgb,255:red,210; green,233; blue,255} , rounded corners = 9.0] (5.875,-4.25) rectangle  node {\huge Budget Aggregation} (15.375,-5.5);
\draw [ fill={rgb,255:red,210; green,233; blue,255} , rounded corners = 9.0] (5.875,-6.75) rectangle  node {\huge DP Mechanism (\textsc{MADLIB})} (15.275,-8);
\draw [ fill={rgb,255:red,255; green,255; blue,128} , rounded corners = 9.0] (7.5,-11.75) rectangle  node {\huge DP Obfuscated Text} (13.75,-13);
\draw [->, >=Stealth] (10.75,2) -- (10.75,0.75);
\draw [->, >=Stealth] (8,-0.5) -- (8,-1.75);
\draw [->, >=Stealth] (13.25,-0.5) -- (13.25,-1.75);
\draw [->, >=Stealth] (8,-3) -- (8,-4.25);
\draw [->, >=Stealth] (13.25,-3) -- (13.25,-4.25);
\draw [->, >=Stealth] (10.75,-5.5) -- (10.75,-6.75);
\draw [->, >=Stealth] (10.75,-8) -- (10.75,-9.25);
\draw [->, >=Stealth] (8.25,13.25) -- (8.25,12);
\draw [->, >=Stealth] (13.25,13.25) -- (13.25,12);
\draw [->, >=Stealth] (5.5,10.75) -- (5.5,9.5);
\draw [->, >=Stealth] (5.5,8.25) -- (5.5,7);
\draw [->, >=Stealth] (15.75,10.75) -- (15.75,9.5);
\draw [->, >=Stealth] (15.75,8.25) -- (15.75,7);
\draw [->, >=Stealth, dashed] (4,5.75) -- (4,-1.75);
\draw [->, >=Stealth, dashed] (21,6.25) .. controls (22.25,5.75) and (22.75,-7.75) .. (15.375,-7.5);
\draw [ fill={rgb,255:red,210; green,233; blue,255} , rounded corners = 9.0] (5.875,-9.25) rectangle  node {\huge Postprocessing} (15.375,-10.5);
\draw [->, >=Stealth] (10.75,-10.5) -- (10.75,-11.75);
\draw  (0,13.5) rectangle (0,13.5);
\node [font=\huge] at (1.75,14.75) {Preparation};
\node [font=\huge] at (1.25,2.75) {Pipeline};
\end{circuitikz}
}%
\caption{Workflow of our systematic evaluation.}
\label{fig:pipeline}
\end{figure}

We outline the steps of our methodology, which span a preparation and privatization pipeline, culminating in an extensive evaluation. The workflow of our methodology is illustrated in Figure \ref{fig:pipeline}.

\subsection{A note on privacy guarantees}
We preface the introduction of text decomposition and privacy budget distribution methods with an important clarification on the \textit{privacy guarantees} offered as a result of our privatization procedures. We measure and evaluate all texts on the \textbf{document level}; by leveraging the basic composition theorem of DP, we can \textit{compose} the individual privatization of decomposed text chunks into a fixed, document-level privacy budget. This is essential to ensure that in evaluation, all texts receive the same document-level privacy budget, regardless of the number of decomposed units. The main goal of privacy budget distribution, therefore, is to allocate this overall budget to the component chunks, in an optimized manner. We detail both decomposition and budget distribution in the following.

\subsection{N-gram Extraction}
As three of our five text decomposition methods (Section \ref{sec:decomp}) rely on association measures, the first step was to extract a corpus of n-grams for score calculation. We used the FineWeb dataset \cite{10.5555/3737916.3738886}, specifically the \textsc{sample-10BT} subset with 10B tokens. The goal of using this dataset was to extract commonly occurring English n-grams, to serve as the basis of text decomposition methods.

We extracted n-grams for $n \in \{1,2,3,4\}$. To create a unified tokenization strategy for n-gram extraction (and for the remainder of this work), we used the simple regex \texttt{\textbackslash b\textbackslash w+\textbackslash b}, using Python's \textsc{re}.

\subsection{Association Measure Calculation}
The next step was to quantify the strength of association between all n-grams (above unigrams), i.e., to detect the most meaningful groups of words. For this, we used three scoring techniques, following recommendations from related work \cite{bhalla-klimcikova-2019-evaluation,gu-etal-2021-comparative}.

\paragraph{Pointwise Mutual Information (PMI).}
PMI measures how often words appear together compared to pure chance \cite{church-hanks-1990-word}. We adopt a similar scheme to \citet{meisenbacher-etal-2024-collocation} for PMI scoring, e.g., for bigram PMI:
\vspace{-3pt}
\[
\text{PMI}(w_1, w_2) = \log_2 \frac{c(w_1, w_2) \cdot N}{c(w_1) \cdot c(w_2)}
\]

\noindent $c$ represent a frequency count from the FineWeb corpus, and $N$ is the total count. We adapt the PMI formula accordingly for trigrams and quad-grams, found in Appendix \ref{sec:assoc}. Following PMI calculation, we kept only n-grams with a frequency $\ge$ 275 and with a $PMI > 2$ (following related work), in order to mitigate bias towards lower frequency n-grams.

\paragraph{Log-Likelihood-Ratio (LLR).}
LLR is a statistical measure similar to PMI that compares observed frequencies of n-grams to their expected frequencies \cite{dunning-1993-accurate}. LLR is calculated from a contingency table of likelihood values, i.e., when two words appear together versus not. LLR is defined for bigrams but can be extended to trigrams and quad-grams (detailed in Appendix \ref{sec:assoc}). We kept the top 5\% of LLR-scored n-grams (for $n > 1$).

\paragraph{t-score.}
The t-score also associates observed counts of n-grams with expected counts \cite{church-hanks-1990-word}, normalized by the standard deviation. The bigram t-score is as follows:
\vspace{-3pt}
\[
t(w_1, w_2) = \frac{c(w_1, w_2)-\frac{c(w_1)\cdot c(w_2)}{N}}{\sqrt{c(w_1,w_2)}}
\]

The adapted formulas for trigrams and quad-grams can be found in Appendix \ref{sec:assoc}. As with LLR, we kept the top 5\% of t-scored n-grams.

\subsection{Text Decomposition}
\label{sec:decomp}
We use five decomposition methods, which take a text as input and return a list of sequential MWEs. 

\paragraph{Association-based Decomposition.}
Using each of the three selected association measures, we designed a greedy decomposition approach, which reads a text from left to right and selects the longest available n-gram match (e.g., \say{all over the world} $\gg$ \say{all over the} $\gg$ \say{all over} $\gg$ \say{all}). This greedy approach was chosen over a score maximization approach, as the latter did not produce significantly different results in early testing. The complete algorithm for association-based decomposition is found in Algorithm \ref{alg:assoc} of the Appendix.

\paragraph{POS-based Decomposition.}
POS-based decomposition segments a text according to defined rules, for example, by combining noun or prepositional phrases. Rather than manually define rules, we trained a \textsc{BigramTagger} from \textsc{nltk} on the CoNLL 2000 shared task data \cite{tjong-kim-sang-buchholz-2000-introduction}, which was focused on text chunking. With this tagger, an input text is decomposed into chunks based on the assigned POS tags, outlined in Algorithm \ref{alg:pos} of the Appendix.

\paragraph{WordNet-based Decomposition.}
The final method uses WordNet lexical database \cite{10.1145/219717.219748} to identify matching n-grams that exist as synsets (entries). As with the associated measures, we read a text left to right and greedily select the longest available matching n-gram that exists in WordNet. For this implementation (Algorithm \ref{alg:wordnet}), we use WordNet from \textsc{nltk} \cite{bird2009natural}.

\paragraph{A Note on Stopwords and Contractions.}
To handle stopwords, we retain all stopwords \textit{within} n-grams (i.e., the second word in a trigram, or second or third in a quad-gram). Otherwise, stopwords (from \textsc{nltk}) are stripped from n-grams, and furthermore, are ignored during privatization.

Similarly, we opted to split contractions, thereby treating them as two words (\say{don} and \say{t}). To remedy this in the text reconstruction phase post-privatization, we recombine pairs of words that match a curated set of common contractions\footnote{\scriptsize\url{https://gist.github.com/J3RN/ed7b420a6ea1d5bd6d06}}.

\paragraph{Embedding Model Training.}
To enable DP text obfuscation \textit{chunk-wise}, an embedding model was trained for each of the five decomposition methods. In this way, a unified embedding model can be created, in which unigrams coexist with bigrams to quad-grams. For this, we train \textsc{word2vec} models \cite{mikolov2013efficientestimationwordrepresentations} using the Gensim\footnote{\scriptsize\url{https://radimrehurek.com/gensim}} implementation with 300-d vectors and all default training values. N-grams are trained together with unigrams by treating n-grams as single words (e.g., \say{all\_over\_the\_world}), thereby capturing relationships between n-grams regardless of \textit{n}. The resulting models served as the basis for DP text obfuscation via embedding perturbation.

\subsection{Privacy Budget Distribution}
Privacy budget distribution assigns each chunk in a decomposed text a fraction of the overall budget ($\varepsilon$), such that the total budget is upheld and optimized. Pseudocode for some of the methods can be found in Algorithms \ref{alg:attention_weights}-\ref{alg:information_content} of the Appendix.

\paragraph{Attention Weights.}
We use attention weights from a \textsc{bert-base-uncased} model \cite{devlin-etal-2019-bert}. After performing a forward pass of an input text through the model, we capture and average attention scores across all 12 attention heads, and again across all 12 layers. Finally, we average the resulting scores across the token dimension to obtain a single scalar score per input. Subword tokens were handled by summing the component scores. In addition, special tokens from \textsc{bert} were filtered out. The final scores were normalized and distributed proportionally across all non-stopwords. 

\paragraph{Integrated Gradients.}
We also use Integrated Gradients \cite{pmlr-v70-sundararajan17a} to capture attention. We first defined a forward function that maps input embeddings to a scalar score, which is simply the sum of the mean-pooled embedding (final hidden state) for each token. Then, using Captum\footnote{\scriptsize\url{https://captum.ai/}}, we calculated attribution vectors for each input token, reaching a final score for budget distribution by using the L2 norm of the vectors. We then filtered out special tokens and normalized the scores, only distributing scores to non-stopwords. 

\paragraph{Information Content.}
Information Content (IC) is a measure of the informational value of a unit of language, and it is calculated based on the frequency of occurrences in a corpus. We utilize pre-calculated IC values from \textsc{nltk}, which contains five variants from different corpora. Since these IC values are tied to WordNet synsets, we first used the \textsc{lesk} function from \textsc{nltk} to disambiguate word sense, and then calculated an IC score by averaging the entry values over all five IC corpora.

\paragraph{\textsc{KeyBERT}.}
\textsc{KeyBERT} \cite{grootendorst2020keybert} is an unsupervised keyword extraction method leveraging BERT-like embedding models. We use \textsc{KeyBERT} with the \textsc{all-MiniLM-L6-v2} model \cite{reimers-gurevych-2019-sentence}, to return a score for all unigrams in an input text (i.e., \textit{top-N} for $N = len(\text{text})$). Since a higher \textsc{KeyBERT} score indicates higher importance as a keyword, we invert these scores for budget distribution.

\paragraph{\textsc{YAKE}.}
We evaluate another unsupervised keyword extraction method, \textsc{YAKE} \cite{campos2020yake}, which is statistical-based rather than transformer-based. As with \textsc{KeyBERT}, we ensure that \textsc{YAKE} returns a score for all unigrams. In contrast, though, lower scores indicate more suitable keyword candidates; thus, scores are left unaltered.

\paragraph{Final Budget Distribution.}
The output of each distribution method is a mapping of individual words from the input text to a corresponding score. To reach a final budget allocation, negative scores were handled by adding the absolute value of the minimum score to all scores. Next, stopword scores were set to 0. If necessary, the remaining scores were inverted. Finally, the scores were normalized to add up to 1, and then scaled by the target $\varepsilon$ budget to fulfill the total budget constraint. Finally, scores belonging to the same decomposed chunk were summed. The steps taken to reach the final distribution and aggregate chunk scores are described in Algorithms \ref{alg:score_conversion} and \ref{alg:chunked_budgets}, respectively.

\section{Experimental Setup}
We design a full factorial experiment, in which all decomposition methods are tested in combination with the selected budget distribution techniques. In addition to the five distribution techniques introduced above, we also test a \say{baseline} method, which is an even distribution of the overall $\varepsilon$ (i.e., $\varepsilon / len(\text{words})$). Thus, a 5x6 factorial experiment is conducted, across two datasets and three privacy budgets, for a total of 180 experimental runs.

\subsection{Datasets}
We use two datasets of user-written texts, which present a plausible case for text obfuscation, where each dataset features a one-to-many mapping of authors to texts, for a constrained set of authors.

\paragraph{Trustpilot Reviews.}
We use a subset of the Trustpilot corpus \cite{10.1145/2736277.2741141}, a large corpus of user reviews. We take a 10k random sample of \textsc{en-US} reviews, which are mapped to either negative sentiment (1-2 stars) or positive sentiment (4-5 stars); neutral reviews are not included. The reviews are also mapped to the gender of the author.

\paragraph{Yelp Reviews.}
Finally, we utilize a dataset of the 10 most-frequently writing authors on the Yelp platform, prepared by \citet{utpala-etal-2023-locally}, taking a 10k random sample. As with Trustpilot, each review is also mapped to a binary sentiment score.

\subsection{Privatization}
Each privatization configuration (\textit{decomposition}, \textit{distribution}, \textit{dataset}) is run on three \textit{privacy levels}, represented by the document-level $\varepsilon$ budget. As introduced, all texts within a dataset are obfuscated with the same $\varepsilon$ level. The exact distribution of this budget among the component chunks of a text is determined by the distribution method in use.

To establish the $\varepsilon$ budgets for the three privacy levels (\textit{high}, \textit{medium}, \textit{low}), we choose base $\varepsilon$ values, namely, $\varepsilon \in \{0.1, 1, 5\}$, respectively. These values were then scaled by the average document length (in words) to achieve the final three document-level budgets per dataset (Table \ref{tab:epsilon_values}). With these, we ensure comparable privacy levels between all privatized texts of a dataset, while also scaling reasonably to the average document length.

\begin{table}[t!]
\centering
\resizebox{0.47\textwidth}{!}{
\begin{tabular}{l|cc}
\toprule
\textbf{Privacy Level} & \textbf{Trustpilot} & \textbf{Yelp} \\
\midrule
\textit{High} ($\epsilon = 0.1 \times \text{avg. doc length}$) & 5.2 & 18.7\\
\textit{Medium} ($\epsilon = 1 \times \text{avg. doc length}$) & 52 & 187\\
\textit{Low} ($\epsilon = 5 \times \text{avg. doc length}$) & 260 & 935\\
\bottomrule
\end{tabular}
}
\caption{Document-level privacy budget ($\epsilon$) values.}
\label{tab:epsilon_values}
\end{table}

Given an input text and a document-level $\varepsilon$, the text is first decomposed into chunks, and then each of these chunks is allocated a portion of the privacy budget via a distribution method. Finally, using the MADLIB method \cite{feyisetan_balle_2020}, an MLDP perturbation mechanism, we obfuscate each of these chunks using the embedding model corresponding to the utilized decomposition method (Section \ref{sec:decomp}). Each chunk is mapped to its embedding in our trained model, calibrated Laplacian noise according to the allocated budget is added to the embedding by MADLIB, and the perturbed embedding is projected back to the nearest embedding. The corresponding n-gram is then inserted as the DP obfuscated output. The output is postprocessed by removing ``\_'' characters that connect n-grams and recombining contractions, where necessary.

\begin{figure*}[t!]
    \centering
    \includegraphics[width=0.99\linewidth]{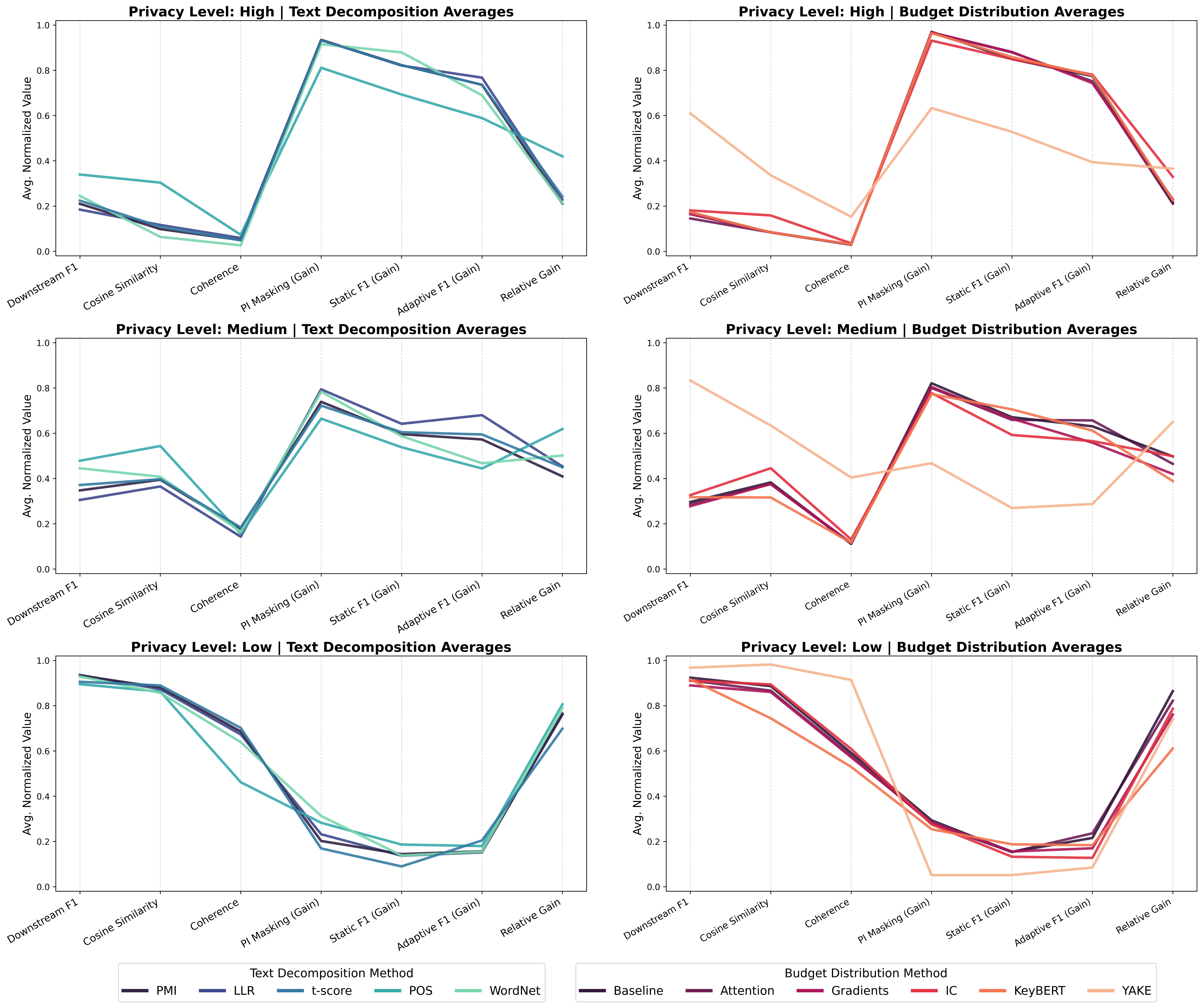}
    \caption{Averaged results over both datasets, for the three selected privacy levels ($\varepsilon$ budgets). The graphs illustrate the average performance of decomposition and budget distribution over the seven captured metrics. For the privacy metrics, the \textit{gain} (i.e., 1 - score) is presented. All scores are normalized between 0-1, with 1 being the highest score.}
    \label{fig:results}
\end{figure*}

\subsection{Evaluation}
Our evaluation takes the form of privacy and utility measurements, as well as the privacy-utility trade-off, which are detailed in the following.

\subsubsection{Privacy Evaluation}
The privacy evaluations analyze two aspects of privacy protection: personal identifier masking and defense against attribute inference.

To enumerate identifiers, we use Microsoft Presidio\footnote{\scriptsize\url{https://microsoft.github.io/presidio/}} to detect all private identifiers (PI) in an input text. Then, we measure what percentage of these identifiers are still (completely) present in the obfuscated output texts. This is averaged over all original-private text pairs, where a lower average score represents better overall privacy protection.

To simulate attribute inference attacks, we adopt the adversarial framework of \textit{static} and \textit{adaptive} attackers \cite{mattern-etal-2022-limits,utpala-etal-2023-locally}. The static attacker is capable of training an adversarial classification on original, non-obfuscated \say{public} texts, thereafter using the trained model to infer sensitive attributes of DP obfuscated texts, i.e., the identity or gender of the original author. The adaptive attacker is further capable of mimicking the DP obfuscation process by first privatizing the training set, training the adversarial model on this data, and then inferring attributes of the target obfuscated texts. For both setups, we train a \textsc{deberta-v3-base} model \cite{he2021debertadecodingenhancedbertdisentangled} for one epoch and use default Hugging Face Trainer parameters, except for the use of the focal loss function, a learning rate of 2e-5, and 500 warmup steps. The training set is a 90\% random split of each dataset, and the target split is the remaining 10\% test set. The resulting score is represented by the adversarial F1 inference performance.

\subsubsection{Utility Evaluation}
The utility evaluations capture three aspects: downstream utility, semantic similarity, and coherence.

For downstream utility, we fine-tune a \textsc{deberta-v3-base} model for one epoch on a 90\% train split of all Trustpilot and Yelp datasets, using the same parameters as the attacker models described above. The F1 score of the trained model on the 10\% test set is reported as an average of three training runs.

To measure semantic similarity between original and obfuscated text counterparts, we measure the average cosine similarity between all pairs in a given dataset. This is performed by using the embeddings calculated by three different pre-trained sentence transformer models, namely \textsc{all-MiniLM-L6-v2} \cite{reimers-gurevych-2019-sentence}, \textsc{all-mpnet-base-v2} (ibid), and \textsc{gte-small} \cite{li2023generaltextembeddingsmultistage}. The resulting scores from the three models are averaged for the final similarity score.

Text coherence can be approximated by \textit{perplexity} to give a sense of the quality of obfuscated text output \cite{10.1145/3485447.3512232,mattern-etal-2022-limits}. We report the average perplexity over the first 32 tokens of all texts in a dataset, calculated using a \textsc{GPT-2} model \cite{radford2019language}.

\subsubsection{Trade-off Calculation}
To represent the privacy-utility trade-off, we calculate the \textit{relative gain} metric, introduced by \citet{mattern-etal-2022-limits}. This metric directly weighs utility losses and privacy gains, and a positive result implies that gained privacy outweighs lost utility. Specifically, we define $RG = \frac{U_p}{U_o} - \frac{P_p}{P_o}$, where $U$ denotes the average utility scores (downstream, similarity), $P$ the average privacy scores (PI, static, adaptive), and the subscripts $_p$ and $_o$ denote the private and original datasets, respectively. Coherence was excluded due to the large/unbounded values.

\section{Results and Statistical Analysis}
The complete results of all 180 evaluation setups are presented in Tables \ref{tab:results_trustpilot_5_2}-\ref{tab:results_trustpilot_935} of Appendix \ref{sec:complete}. In Figure \ref{fig:results}, we illustrate the aggregated (average) results over both evaluation datasets, for all three privacy levels and for both decomposition and distribution.

Figure \ref{fig:results} clearly shows that despite equal document-level $\varepsilon$ budgets, different results can be obtained for privacy, utility, and relative gains. Taking the relative gain (trade-off) as a dependent variable, we conduct significance tests to determine if the choice of decomposition or distribution method affects the resulting trade-offs. For this, a two-way ANOVA test \cite{Fisher1992} is fitting, as we are studying two categorical independent variables (decomposition and distribution) and one continuous dependent variable (relative gain). We also conduct one-way tests on the effect of decomposition \textit{or} distribution. All tests are performed using \textsc{statsmodel}. The categorical variables \textit{dataset} and \textit{privacy level} are included as controls.

We find that the choice of both decomposition ($F = 5.57, p < 0.001$) and distribution ($F = 3.97, p = 0.002$) has significant effects on the resulting relative gain; however, the \textit{interaction} between the two, as concluded by the two-way test, is not significant ($F = 0.63, p = 0.88$). While the choice of distribution method provides the largest absolute relative gain gap (0.0324 vs. 0.0290) over decomposition, the higher $F$-statistic implies that decomposition provides a more consistent impact on relative gain. This is confirmed by a partial eta squared (effect size) value of $\eta_p^2 = 0.132$ for decomposition and $\eta_p^2 = 0.119$ for distribution.

\begin{figure}[t]
    \centering
    \includegraphics[width=0.95\linewidth]{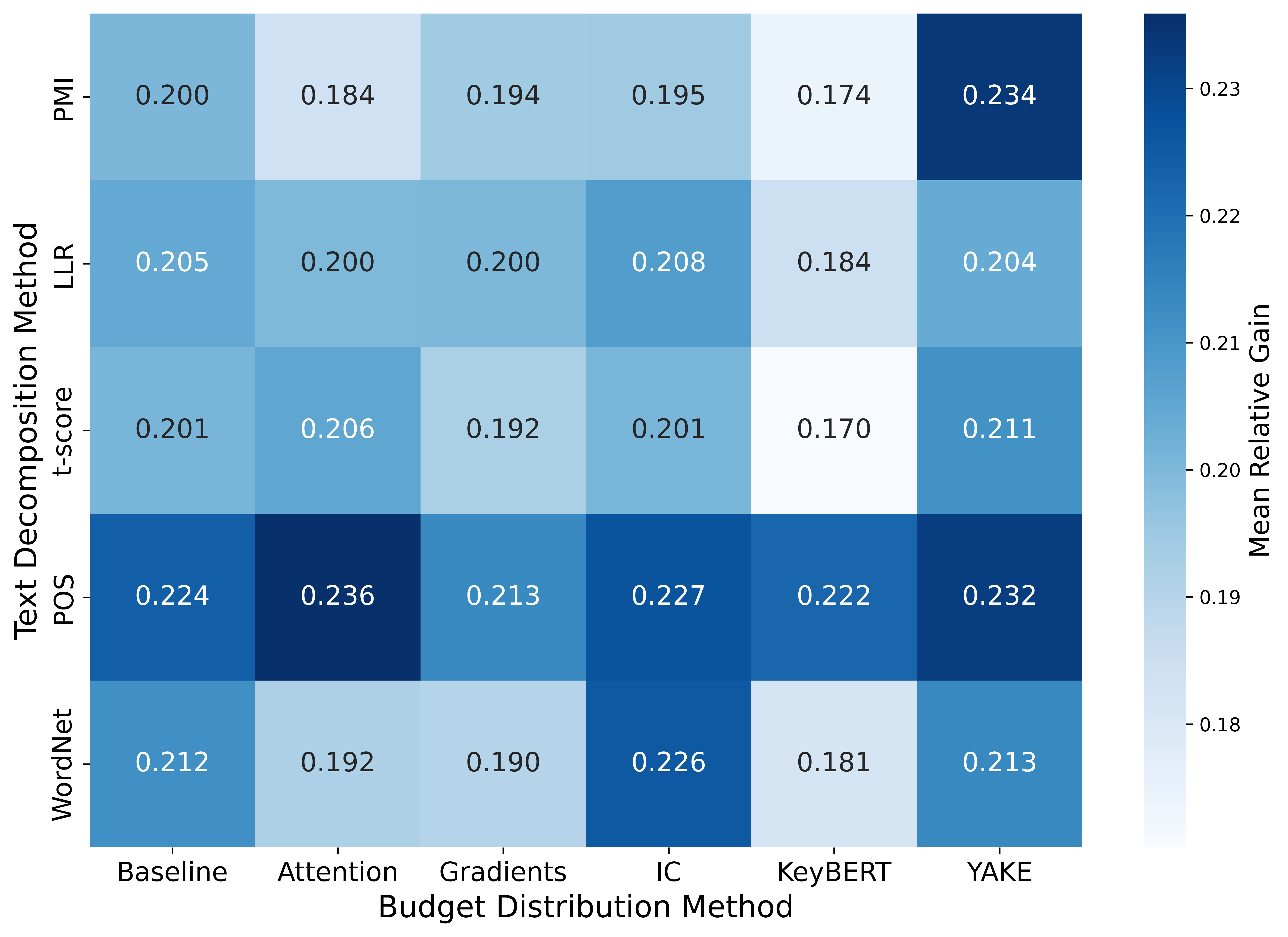}
    \caption{Global relative gain averages ($\uparrow$), i.e., over two datasets and three privacy levels.}
    \label{fig:heatmap}
    \vspace{-5pt}
\end{figure}

To investigate further, we normalize all scores by the mean of each (\textit{dataset}, \textit{privacy level}) group to remove confounding effects. Then, we perform a Tukey's Honestly Significant Difference (HSD) post-hoc test \cite{tukey}, which tests for significance between all decomposition-distribution pairs (435 in total, or $\frac{(30\times30)}{2} - \frac{30}{2}$). The HSD analysis reveals significant differences ($p < 0.05$) between \textbf{25} pairs. Notably, the combination of POS+attention achieved the largest significant differences (in magnitude), followed by PMI+YAKE and POS+YAKE. Conversely, 21 of these 25 differences feature KeyBERT in the lower end of the comparison. These results are supported by an illustration of the global averages for relative gain, depicted in Figure \ref{fig:heatmap}.

\section{Discussion}
We reflect on the main findings of our systematic evaluation and discuss their implications.

\paragraph{Break it down and distribute.}
The findings from our experiments and the resulting analysis demonstrate the importance of selecting appropriate methods for text decomposition and privacy budget distribution in the context of DP text obfuscation. We discover that these decisions can lead to significant differences in the downstream trade-offs, despite using exactly the same document-level $\varepsilon$ budgets over a dataset. Interestingly, while the choice of decomposition and distribution is individually significant, their \textit{interaction} is not (by far), suggesting that there are no conflicting effects between the two stages of DP text obfuscation. 

These insights support the importance of informed DP text obfuscation, where \say{simple} uniform budget distribution rarely produces superior results, and equally as importantly, where the manner in which text documents are chunked is crucial to maximizing utility and privacy. Thus, we provide evidence of the fact that equivalent theoretical privacy guarantees (via $\varepsilon$) does not necessitate equivalent empirical results, and this must be carefully tuned to optimize privacy-utility trade-offs.

\paragraph{What works well together?}
Looking into potential optimal combinations of decomposition and distribution techniques, as found in our experiment results, reveals interesting insights. Optimizing for relative gains yields the clear choice of POS+Attention, a pair which achieved an average trade-off of 0.236. The combination of PMI+YAKE is the best-performing strategy for downstream utility preservation. When considering all three utility metrics (F1, CS, and Perplexity), the top-five results (averaged across all setups) \textit{always} feature YAKE as the distribution method.

In scenarios where privacy protection is crucial, though, different winners emerge. Averaging the three privacy metrics we employ, the combination of LLR+\textsc{KeyBERT} achieves the best results across all setups. In the top-five average privacy results, LLR and WordNet both appear twice for decomposition, and likewise for \textsc{KeyBERT} and Attention. Thus, while our statistical analysis shows that choice of decomposition and distribution can be considered separately, our results imply that when optimizing for privacy, utility, and their trade-offs, particular strategies may be more well-suited.

\begin{table}[t]
\resizebox{0.99\linewidth}{!}{
\begin{tabular}{l|cccc}
 & sum\_sq & df & F & $p$ \\ \hline
C(Decomposition) & 0.0214 & 4.0 & 5.57 & \textbf{3.32e-04} \\
C(Distribution) & 0.0191 & 5.0 & 3.97 & \textbf{2.10e-03} \\
C(Dataset) & 0.0370 & 1.0 & 38.44 & \textbf{5.42e-09} \\
Q(Privacy Level) & 0.3173 & 2.0 & 165.01 & \textbf{2.67e-38} \\
C(Decomposition):C(Distribution) & 0.0122 & 20.0 & 0.63 & 8.83e-01 \\
Residual & 0.1414 & 147.0 & -- & --
\end{tabular}
}
\caption{ANOVA test results, with the sum of squares, degrees of freedom (df), the \textit{F}-statistic, and the p-value of \textit{F}. Significant $p$-values ($p$ \textless \ 0.05) are \textbf{bolded}.}
\label{tab:anova}
\end{table}

\paragraph{The curious case of trade-offs.}
Looking purely at resulting trade-offs, though, we learn that this metric is not only significantly influenced by decomposition and distribution, but also by dataset and privacy level effects (as evidenced in the full analysis results in Table \ref{tab:anova}). The impact of \textit{privacy level}, or $\varepsilon$ budget, is also made clear in Figure \ref{fig:results}, which paradoxically illustrates average relative gains increasing as the privacy level decreases. 

This \say{curious case} calls to question the possible dominance of utility measurement in privacy-utility trade-off calculations. More importantly, it points to the idea that there presumably exists a (local) maximum in terms of trade-offs; however, this must be meticulously balanced such that the boundary case of very high utility preservation and very low privacy gains does not become the optimal result.

\paragraph{Practical implications.}
The results of our systematic evaluation carry practical implications. While our experiments are limited in scope and do not represent a comprehensive factorial analysis of decomposition and distribution in DP text obfuscation, it does define a blueprint for doing so at scale. We show that with the support of an experiment setup like ours, practitioners can become informed on which particular setups work optimally, for a given data domain and privacy preference. This becomes important for providing explainability and usability to the scalable $\varepsilon$ parameter, and it allows for flexibility in defining priority objectives, which we assume in this work to be positive trade-offs.

We highlight the need to view the integration of DP and textual data as a \textit{linguistics}-inspired task. This embodies \textit{divide and conquer}, working with text in tandem with considerations of optimal groupings of semantic meaning, as well as the relative quantification of their \say{importance} in context. With this \textit{divide and conquer} mindset, we argue that DP text privatization can not only be optimized, but also be more aligned to true privacy protection.

\section{Conclusion}
We conduct a systematic evaluation of five text decomposition methods and six privacy budget distribution methods for document-level DP text obfuscation. Our experiments demonstrate the importance of the design of privatization procedures, where the precise allocation of the privacy budget has significant implications on downstream privacy and utility. As such, we advance the understanding and optimization of DP text privatization, providing a foundation for future work on (1) designing and evaluating intelligent methods for decomposing texts into coherent chunks, (2) devising methods for quantifying the \say{importance} of such chunks to utility and privacy, and (3) conducting systematic studies on the factors contributing to trade-offs when operating local DP mechanisms on text.

\section*{Limitations}
Our work is limited by the selected methods for text decomposition and privacy budget distribution, which we selected to be a representative sample of suitable methods from an informal literature review. Future work could build on this foundation to explore a wide breadth of methods, which would also serve to validate our findings.

Since the design and evaluation of DP mechanisms was not central to our work, we only used the original MADLIB implementation from \citet{feyisetan_balle_2020}, adapted to function with our various trained embedding models. As such, we cannot make any conclusions regarding the (statistical) effect of mechanism choice, which could be added in future systematic evaluations for a more complete picture of DP text obfuscation effectiveness.

Limitations to our evaluation procedure also include the sole focus on the English language, as well as the limited domain scope (user review texts). Follow-up studies should expand evaluation to other languages and domains for greater generalizability of our findings.

As with any quantification of privacy, utility, and the privacy-utility trade-off, our evaluations are limited by the chosen metrics, which provide a snapshot of what one may consider privacy or utility. Particularly with privacy evaluation, the measure of personal identifier masking and attribute inference detection does not constitute a holistic test of privacy preservation; we use these metrics as a proxy, following previous works in the field.

\section*{Ethical Considerations}
We confirm that all utilized software and datasets for our experiments are open-source and acceptable to use under their respective licenses. No data was collected from human subjects.


\bibliography{custom}

\appendix

\section{Reproducibility}

\paragraph{Hardware.}
All CPU-based experiments were run on a single Intel Xeon Gold 6148 20-core CPU, and all experiments benefiting from GPU acceleration were run on a Nvidia Tesla V100 16GB GPU.

\paragraph{Random Seeding.}
For any procedure requiring randomization, we used a random seed of 42. For training tasks where we ran three training runs (and reported the average score), we used the seeds of 42, 43, and 44, sequentially.

\paragraph{Training Parameters.}
The training parameters used for \textsc{word2vec} models are given in Table \ref{tab:word2vec_params}.

For the privacy and utility experiments which involved the fine-tuning of a \textsc{deberta-v3-base}, the training parameters are reported in Table \ref{tab:params}.

\begin{table}[ht!]
\centering
\resizebox{0.7\linewidth}{!}{
\begin{tabular}{lll}
\toprule
\textbf{Parameter} & \textbf{Value} \\
\midrule
Vector Size & 300 \\
Architecture & Skip-gram \\
Training Workers & 5 \\
Window Size & 5 \\
Minimum Count & 5 \\
Training Epochs & 5 \\
Negative Samples & 5 \\
\bottomrule
\end{tabular}
}
\caption{Word2Vec Training Parameters}
\label{tab:word2vec_params}
\end{table}

\begin{table}[ht!]
\centering
\resizebox{0.8\linewidth}{!}{
\begin{tabular}{ll}
\toprule
\textbf{Parameter} & \textbf{Value} \\
\midrule
Model & \textsc{deberta-v3-base} \\
Num. Epochs & 1 \\
Learning Rate & 2e-5 \\
Warmup Steps & 500 \\
Batch Size & 8 \\
Loss Function & Focal Loss \\
\quad Focal Loss $\gamma$ & 2.0 \\
\quad Focal Loss $\alpha$ & 0.25 \\
Train/Test Split & 90\% / 10\% \\
Number of Runs & 3 \\
Random Seed & 42 \\
\bottomrule
\end{tabular}
}
\caption{\textsc{DeBERTa} Training Parameters.}
\label{tab:params}
\end{table}

\section{Association Measure Calculation}
\label{sec:assoc}
With $N$ as the total unigram count and $c$ as a frequency count, the trigram PMI score of $(w_1, w_2, w_3)$ is defined as:
\small
\[
\text{PMI}(w_1, w_2, w_3) = \log_2 \frac{c(w_1, w_2, w_3) \cdot N^2}{c(w_1) \cdot c(w_2) \cdot c(w_3)}
\]
\normalsize

Similarly, the quad-gram PMI score:
\small
\[
\text{PMI}(w_1, w_2, w_3, w_4) = \log_2 \frac{c(w_1, w_2, w_3, w_4) \cdot N^3}{c(w_1) \cdot c(w_2) \cdot c(w_3) \cdot c(w_4)}
\]
\normalsize

For LLR, the calculated scores are based on a 2x2 contingency table (Table \ref{tab:llr}), exemplified by:
\tiny
\[
\begin{split}
\text{LLR} = 2 \Biggl( & x\log x(N) - x\log x(c_{11} + c_{12}) - x\log x(c_{21} + c_{22}) \\
                     & - x\log x(c_{11} + c_{21}) - x\log x(c_{12} + c_{22}) \\
                     & + x\log x(c_{11}) + x\log x(c_{12}) + x\log x(c_{21}) + x\log x(c_{22}) \Biggr)
\end{split}
\]
\normalsize

\begin{table}[t]
\centering
\resizebox{0.9\linewidth}{!}{
\begin{tabular}{c|c|c}
\toprule
 & \textbf{$w_1$} & \textbf{Not $w_1$} \\
\midrule
\textbf{$w_2$} & c($w_1, w_2$) &  c($w_2$ not $w_1$)\\
\textbf{Not $w_2$} & c($w_1$ not $w_2$) & c(not $w_1$ or $w_2$)\\
\bottomrule
\end{tabular}
}
\caption{Example Contingency Table for LLR Score.}
\label{tab:llr}
\end{table}

Finally, we define the trigram t-score as:
\small
\[
t(w_1, w_2, w_3) = \frac{c(w_1, w_2, w_3) - \frac{c(w_1) \cdot c(w_2) \cdot c(w_3)}{N^2}}{\sqrt{c(w_1, w_2, w_3)}}
\]
\normalsize

And the quad-gram t-score as:
\scriptsize
\[
t(w_1, w_2, w_3, w_4) = \frac{c(w_1, w_2, w_3, w_4) - \frac{c(w_1) \cdot c(w_2) \cdot c(w_3) \cdot c(w_4)}{N^3}}{\sqrt{c(w_1, w_2, w_3, w_4)}}
\]
\normalsize

\section{Algorithm Pseudocode}
\label{sec:algorithm}
The algorithms for text decomposition are outlined in Algorithm \ref{alg:assoc} for association-based methods, Algorithm \ref{alg:pos} for POS-based decomposition, and Algorithm \ref{alg:wordnet} for WordNet-based decomposition.

Detailed pseudocode for budget distribution methods can be found for Attention Weights (Algorithm \ref{alg:attention_weights}), Integrated Gradients (Algorithm \ref{alg:integrated_gradients}), and Information Content (Algorithm \ref{alg:information_content}). 

\begin{algorithm}[ht]
\scriptsize
\caption{\small Association-based Decomposition (PMI/LLR/t-score)}
\label{alg:assoc}
\KwIn{document $D$, n-gram sets $\mathcal{B}$, $\mathcal{T}$, $\mathcal{Q}$, stopwords $\mathcal{S}$}
\KwOut{chunk sequence $C$}
\BlankLine

$C \gets [\,]$\;
$\text{sentences} \gets \text{SentenceTokenize}(D)$\;
\BlankLine

\ForEach{$sentence \in \text{sentences}$}{
    $\text{tokens} \gets \text{WordTokenize}(sentence, \text{pattern}=\backslash\text{b}\backslash\text{w}+\backslash\text{b})$\;
    $\text{tokens} \gets \text{Lowercase}(\text{tokens})$\;
    $i \gets 0$\;
    \BlankLine
    
    \While{$i < |\text{tokens}|$}{
        $\text{matched} \gets \text{False}$\;
        \BlankLine
        
        \If{$i + 3 < |\text{tokens}|$}{
            $\text{ngram} \gets \text{Join}(\text{tokens}[i:i+4], \text{" "})$\;
            \If{$\text{ngram} \in \mathcal{Q}$}{
                $C.\text{extend}(\text{ProcessStopwords}(\text{tokens}[i:i+4], \mathcal{S}))$\;
                $i \gets i + 4$\;
                $\text{matched} \gets \text{True}$\;
                \textbf{continue}\;
            }
        }
        \BlankLine
        
        \If{$\neg\text{matched} \land i + 2 < |\text{tokens}|$}{
            $\text{ngram} \gets \text{Join}(\text{tokens}[i:i+3], \text{" "})$\;
            \If{$\text{ngram} \in \mathcal{T}$}{
                $C.\text{extend}(\text{ProcessStopwords}(\text{tokens}[i:i+3], \mathcal{S}))$\;
                $i \gets i + 3$\;
                $\text{matched} \gets \text{True}$\;
                \textbf{continue}\;
            }
        }
        \BlankLine
        
        \If{$\neg\text{matched} \land i + 1 < |\text{tokens}|$}{
            $\text{ngram} \gets \text{Join}(\text{tokens}[i:i+2], \text{" "})$\;
            \If{$\text{ngram} \in \mathcal{B}$}{
                $C.\text{extend}(\text{ProcessStopwords}(\text{tokens}[i:i+2], \mathcal{S}))$\;
                $i \gets i + 2$\;
                $\text{matched} \gets \text{True}$\;
                \textbf{continue}\;
            }
        }
        \BlankLine
        
        $C.\text{append}(\text{tokens}[i])$\;
        $i \gets i + 1$\;
    }
}
\BlankLine

$C \gets \text{MergeContractions}(C)$\;
\Return{$C$}\;
\end{algorithm}

\begin{algorithm}[p]
\scriptsize
\caption{POS-based Decomposition with BigramTagger}
\label{alg:pos}
\KwIn{document $D$, trained BigramTagger $\theta$, stopwords $\mathcal{S}$}
\KwOut{chunk sequence $C$}
\BlankLine

$C \gets [\,]$\;
$\text{sentences} \gets \text{SentenceTokenize}(D)$\;
\BlankLine

\ForEach{$sentence \in \text{sentences}$}{
    $\text{tokens} \gets \text{WordTokenize}(sentence)$\;
    $\text{tokens} \gets \text{Lowercase}(\text{tokens})$\;
    \BlankLine
    
    $\text{pos\_tags} \gets \text{POSTagger}(\text{tokens})$\;
    \BlankLine
    
    $\text{chunk\_tags} \gets \theta.\text{tag}(\text{pos\_tags})$\;
    \BlankLine
    
    $\text{tree} \gets \text{ParseIOB}(\text{tokens}, \text{pos\_tags}, \text{chunk\_tags})$\;
    \BlankLine
    
    \ForEach{$\text{subtree} \in \text{tree}$}{
        \eIf{$\text{IsChunk}(\text{subtree})$}{
            $\text{words} \gets \text{ExtractWords}(\text{subtree})$\;
            $\text{processed} \gets \text{ProcessStopwords}(\text{words}, \mathcal{S})$\;
            $C.\text{extend}(\text{processed})$\;
        }{
            $\text{word} \gets \text{ExtractWord}(\text{subtree})$\;
            $C.\text{append}(\text{word})$\;
        }
    }
}
\BlankLine

$C \gets \text{MergeContractions}(C)$\;
\Return{$C$}\;
\end{algorithm}

\begin{algorithm}[p]
\scriptsize
\caption{WordNet-based Decomposition}
\label{alg:wordnet}
\KwIn{document $D$, WordNet database $\mathcal{W}$, stopwords $\mathcal{S}$}
\KwOut{chunk sequence $C$}
\BlankLine

$C \gets [\,]$\;
$\text{sentences} \gets \text{SentenceTokenize}(D)$\;
\BlankLine

\ForEach{$sentence \in \text{sentences}$}{
    $\text{tokens} \gets \text{WordTokenize}(sentence)$\;
    $\text{tokens} \gets \text{Lowercase}(\text{tokens})$\;
    $i \gets 0$\;
    \BlankLine
    
    \While{$i < |\text{tokens}|$}{
        $\text{matched} \gets \text{False}$\;
        \BlankLine
        
        \If{$i + 3 < |\text{tokens}|$}{
            $\text{lemma} \gets \text{Join}(\text{tokens}[i:i+4], \text{"\_"})$\;
            \If{$\mathcal{W}.\text{synsets}(\text{lemma}) \neq \emptyset$}{
                $C.\text{extend}(\text{ProcessStopwords}(\text{tokens}[i:i+4], \mathcal{S}))$\;
                $i \gets i + 4$\;
                $\text{matched} \gets \text{True}$\;
                \textbf{continue}\;
            }
        }
        \BlankLine
        
        \If{$\neg\text{matched} \land i + 2 < |\text{tokens}|$}{
            $\text{lemma} \gets \text{Join}(\text{tokens}[i:i+3], \text{"\_"})$\;
            \If{$\mathcal{W}.\text{synsets}(\text{lemma}) \neq \emptyset$}{
                $C.\text{extend}(\text{ProcessStopwords}(\text{tokens}[i:i+3], \mathcal{S}))$\;
                $i \gets i + 3$\;
                $\text{matched} \gets \text{True}$\;
                \textbf{continue}\;
            }
        }
        \BlankLine
        
        \If{$\neg\text{matched} \land i + 1 < |\text{tokens}|$}{
            $\text{lemma} \gets \text{Join}(\text{tokens}[i:i+2], \text{"\_"})$\;
            \If{$\mathcal{W}.\text{synsets}(\text{lemma}) \neq \emptyset$}{
                $C.\text{extend}(\text{ProcessStopwords}(\text{tokens}[i:i+2], \mathcal{S}))$\;
                $i \gets i + 2$\;
                $\text{matched} \gets \text{True}$\;
                \textbf{continue}\;
            }
        }
        \BlankLine
        
        $C.\text{append}(\text{tokens}[i])$\;
        $i \gets i + 1$\;
    }
}
\BlankLine

$C \gets \text{MergeContractions}(C)$\;
\Return{$C$}\;
\end{algorithm}

\begin{algorithm}[p]
\scriptsize
\caption{Extract Attention Weights}
\label{alg:attention_weights}
\KwIn{text $T$, privacy budget $\epsilon$}
\KwOut{token-budget pairs $(w_i, \epsilon_i)$}
\BlankLine
$\text{tokens} \gets \text{BertTokenizer}(T)$\;
$\text{inputs} \gets \text{prepare\_inputs}(\text{tokens})$\;
$\text{outputs} \gets \text{BertModel}(\text{inputs})$\;
$A \gets \text{outputs.attentions}$\;
\BlankLine
$A_{\text{avg}} \gets \text{mean}(A, \text{dim}=(\text{layers}, \text{heads}))$\;
$\text{scores} \gets \text{mean}(A_{\text{avg}}, \text{dim}=\text{tokens})$\;
\BlankLine
$\text{scored\_tokens} \gets \text{zip}(\text{tokens}, \text{scores})$\;
$\text{combined} \gets \text{combine\_subwords}(\text{scored\_tokens})$\;
$\text{filtered} \gets \text{remove}(\text{combined}, \text{special\_tokens})$\;
\BlankLine
$\text{budgets} \gets \text{normalize\_and\_distribute}(\text{filtered}, \epsilon)$\;
\Return{$\text{budgets}$}\;
\end{algorithm}

\begin{algorithm}[p]
\scriptsize
\caption{Extract Integrated Gradients}
\label{alg:integrated_gradients}
\KwIn{text $T$, privacy budget $\epsilon$}
\KwOut{token-budget pairs $(w_i, \epsilon_i)$}
\BlankLine
$\text{tokens} \gets \text{BertTokenizer}(T)$\;
$E \gets \text{BertEmbeddings}(\text{tokens})$\;
\BlankLine
\SetKwFunction{Forward}{ForwardFunc}
\SetKwProg{Fn}{Function}{:}{}
\Fn{\Forward{$E$}}{
    $H \gets \text{BertModel}(\text{inputs\_embeds}=E)$\;
    \Return{$\text{mean}(H).sum()$}\;
}
\BlankLine
$\text{attributions} \gets \text{IntegratedGradients}(\Forward, E)$\;
$\text{scores} \gets ||\text{attributions}||_2$\;
\BlankLine
$\text{scored\_tokens} \gets \text{zip}(\text{tokens}, \text{scores})$\;
$\text{combined} \gets \text{combine\_subwords}(\text{scored\_tokens})$\;
$\text{filtered} \gets \text{remove}(\text{combined}, \text{special\_tokens})$\;
\BlankLine
$\text{budgets} \gets \text{normalize\_and\_distribute}(\text{filtered}, \epsilon)$\;
\Return{$\text{budgets}$}\;
\end{algorithm}

\begin{algorithm}[p]
\scriptsize
\caption{Get Information Content}
\label{alg:information_content}
\KwIn{text $T$, privacy budget $\epsilon$}
\KwOut{token-budget pairs $(w_i, \epsilon_i)$}
\BlankLine
$\text{tokens} \gets \text{tokenize}(T)$\;
$\text{pos\_tags} \gets \text{pos\_tag}(\text{tokens})$\;
\BlankLine
\ForEach{$(w_i, \text{pos}_i) \in \text{pos\_tags}$}{
    $\text{wn\_pos} \gets \text{convert\_to\_wordnet}(\text{pos}_i)$\;
    $\text{sense} \gets \text{Lesk}(\text{tokens}, w_i, \text{wn\_pos})$\;
    \BlankLine
    \eIf{$\text{sense} \neq \text{null}$}{
        $\text{ic\_values} \gets [\,]$\;
        \ForEach{$\text{corpus} \in \text{IC\_corpora}$}{
            $\text{ic\_values}.\text{append}(\text{IC}(\text{sense}, \text{corpus}))$\;
        }
        $\text{score}_i \gets \text{mean}(\text{ic\_values})$\;
    }{
        $\text{score}_i \gets 1.0$\;
    }
}
\BlankLine
$\text{filtered} \gets \text{filter\_alphanumeric}(\text{scored\_tokens})$\;
$\text{budgets} \gets \text{normalize\_and\_distribute}(\text{filtered}, \epsilon)$\;
\Return{$\text{budgets}$}\;
\end{algorithm}

\begin{algorithm}[p]
\scriptsize
\caption{Convert Scores to Budget Distribution}
\label{alg:score_conversion}
\KwIn{scored tokens $(t_i, s_i)$, original text $T$, privacy budget $\epsilon$, invert flag}
\KwOut{token-budget pairs $(w_i, \epsilon_i)$}
\BlankLine

$\text{original\_words} \gets \text{tokenize}(T)$\;
$\text{score\_map} \gets \text{dict}(\text{scored\_tokens})$\;
\BlankLine

$\text{scores} \gets [\,]$\;
\ForEach{$word \in \text{original\_words}$}{
    \eIf{$word \in \text{stopwords}$}{
        $\text{scores}.\text{append}(0.0)$\;
    }{
        $score \gets \text{score\_map}.\text{get}(word, 0.0)$\;
        $\text{scores}.\text{append}(score)$\;
    }
}
\BlankLine

\If{$\exists s \in \text{scores} : s < 0$}{
    $\text{min\_val} \gets \min(\{s \mid s \neq 0\})$\;
    \ForEach{$s_i \in \text{scores}$ where $s_i \neq 0$}{
        $s_i \gets s_i + |min\_val|$\;
    }
}
\BlankLine

\eIf{$invert = \text{True}$}{
    $\text{non\_zero\_scores} \gets \{s \mid s > 0\}$\;
    $\text{min\_nz} \gets \min(\text{non\_zero\_scores})$\;
    $\text{max\_nz} \gets \max(\text{non\_zero\_scores})$\;
    \BlankLine
    \eIf{$\text{max\_nz} = \text{min\_nz}$}{
        $\text{inverted} \gets \text{ones}(\text{length}(\text{non\_zero\_scores}))$\;
    }{
        $\text{inverted} \gets (\text{max\_nz} + \text{min\_nz}) - \text{non\_zero\_scores}$\;
    }
    $\text{budgets} \gets (\text{inverted} / \sum \text{inverted}) \times \epsilon$\;
}{
    $\text{total} \gets \sum \text{scores}$\;
    \If{$\text{total} > 0$}{
        $\text{budgets} \gets (\text{scores} / \text{total}) \times \epsilon$\;
    }
}
\BlankLine

\Return{$\text{zip}(\text{original\_words}, \text{budgets})$}\;
\end{algorithm}

\begin{algorithm}[p]
\scriptsize
\caption{Get Chunked Budgets}
\label{alg:chunked_budgets}
\KwIn{clean text $T_{clean}$, chunker, distributor, privacy budget $\epsilon$}
\KwOut{chunk-budget pairs $(c_i, \epsilon_i)$, chunks list $C$}
\BlankLine

$\text{chunked\_sentences} \gets \text{chunker}.\text{chunk}(T_{clean})$\;
$C \gets \text{flatten}(\text{chunked\_sentences})$\;
\BlankLine

$\text{word\_budgets} \gets \text{distributor}.\text{distribute}(T_{clean}, \epsilon)$\;
$\text{budget\_iter} \gets \text{iterator}(\text{word\_budgets})$\;
$\text{chunk\_budgets} \gets [\,]$\;
\BlankLine

\ForEach{$chunk \in C$}{
    $\text{words} \gets \text{split}(chunk, \text{`\_'})$\;
    $n \gets |\text{words}|$\;
    $\epsilon_{chunk} \gets 0.0$\;
    \BlankLine
    \For{$i \gets 1$ \KwTo $n$}{
        $(w, \epsilon_w) \gets \text{next}(\text{budget\_iter})$\;
        $\epsilon_{chunk} \gets \epsilon_{chunk} + \epsilon_w$\;
    }
    \BlankLine
    $\text{chunk\_budgets}.\text{append}((chunk, \epsilon_{chunk}))$\;
}
\BlankLine

\Return{$\text{chunk\_budgets}, C$}\;
\end{algorithm}

\section{Complete Results}
\label{sec:complete}
\begin{figure}[p]
    \centering
    \includegraphics[width=0.99\linewidth]{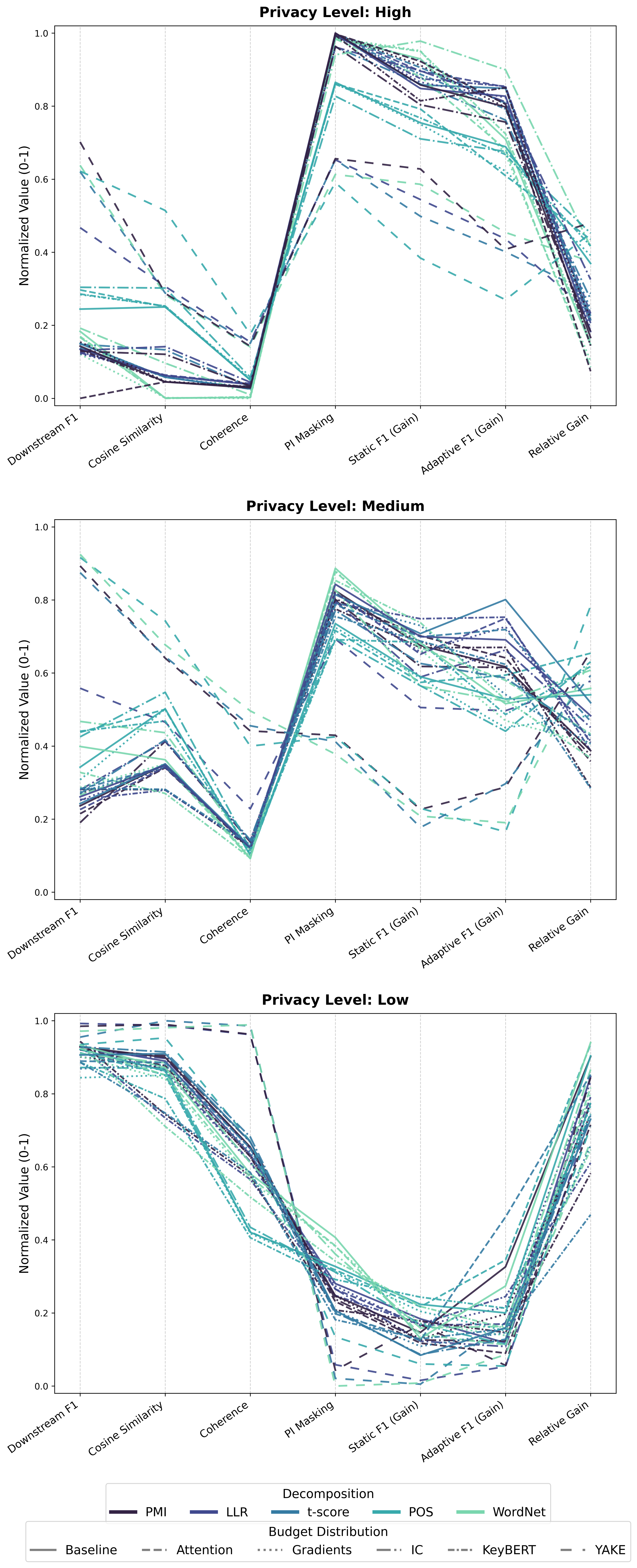}
    \caption{Averaged experiment results over the two selected datasets. Each line represents the mean results for a decomposition-distribution pair, and the y axes plot the normalized scores for all captured metrics. For all privacy metrics, the \textit{gain} (1 - score) is shown, such that for all metrics, 1 represents the best score.}
    \label{fig:results_full}
\end{figure}

The complete experiment results from Trustpilot with privacy levels of high, medium, and low are presented in Tables \ref{tab:results_trustpilot_5_2}, \ref{tab:results_trustpilot_52}, and \ref{tab:results_trustpilot_260}, respectively. Likewise, the complete results from Yelp are presented in Tables \ref{tab:results_trustpilot_18_7}, \ref{tab:results_trustpilot_187}, and \ref{tab:results_trustpilot_935}. As Figure \ref{fig:results} only presents the average scores for both decomposition and distribution, we visualize the averaged results for all 30 combinations in Figure \ref{fig:results_full}.

\begin{table*}[htbp]
\centering
\resizebox{\textwidth}{!}{%
\begin{tabular}{l|l|cccccc|c}
\hline
\multicolumn{2}{c|}{\multirow{2}{*}{\begin{tabular}{c}\textbf{Trustpilot} \\ $\varepsilon=5.2$\end{tabular}}} & \multicolumn{6}{c|}{\textbf{Budget Distribution}} & \multirow{3}{*}{\begin{tabular}{c}\textbf{Average}\end{tabular}} \\
\cline{3-8}
\multicolumn{2}{c|}{}  & \textbf{Baseline} & \textbf{Attention} & \textbf{Gradients} & \textbf{IC} & \textbf{KeyBERT} & \textbf{YAKE} &  \\
\cline{1-8}
\textbf{Decomposition} & \textbf{Baseline} & \multicolumn{6}{c|}{Utility: $98.2_{0.4}$ / Adversary: $72.1_{1.0}$} &  \\
\hline
\multirow{7}{*}{\textbf{PMI}} & Utility F1 & $76.1_{2.3}$ & $67.2_{16.7}$ & $76.8_{2.8}$ & $75.2_{2.1}$ & $75.5_{5.5}$ & $88.6_{0.9}$ & 76.6 \\
& CS & 0.364 & 0.363 & 0.365 & 0.421 & 0.364 & 0.425 & 0.384  \\
& Perplexity & 1497 & 1506 & 1499 & 1414 & 1492 & 1099 & 1418 \\ \cline{2-9} 
&  PI Masking & 0.76 & 0.80 & 0.78 & 2.88 & 0.95 & 8.27 & 2.41 \\
& Static F1 & $43.6_{4.3}$ & $39.0_{0.5}$ & $40.0_{1.4}$ & $47.1_{5.5}$ & $47.0_{5.9}$ & $41.1_{3.5}$ & 43.0 \\
& Adaptive F1 & $55.5_{3.8}$ & $56.6_{1.7}$ & $56.3_{1.2}$ & $57.2_{2.3}$ & $53.7_{2.0}$ & $57.9_{1.2}$ & 56.2 \\ \cline{2-9} 
& Relative Gain & 0.159 & 0.127 & 0.175 & 0.153 & 0.149 & $\mathbf{0.223_2}$ & 0.164 \\
\hline \hline
\multirow{7}{*}{\textbf{LLR}} & Utility F1 & $75.6_{0.3}$ & $75.0_{0.5}$ & $74.8_{3.6}$ & $75.4_{0.9}$ & $75.1_{1.4}$ & $88.4_{0.8}$ & 77.4 \\
& CS & 0.383 & 0.384 & 0.385 & 0.442 & 0.384 & 0.443 & 0.404 \\
& Perplexity & 1348 & 1356 & 1345 & 1273 & 1325 & 1010 & 1276 \\ \cline{2-9} 
&  PI Masking & 0.79 & 0.84 & 0.93 & 2.96 & 1.03 & 8.31 & 2.48 \\
& Static F1 & $45.1_{4.1}$ & $42.1_{3.9}$ & $44.0_{5.2}$ & $41.6_{2.8}$ & $43.0_{4.9}$ & $47.5_{6.0}$ & 43.9 \\
& Adaptive F1 & $56.7_{2.2}$ & $55.1_{0.8}$ & $55.9_{1.9}$ & $55.5_{2.9}$ & $55.7_{1.3}$ & $58.3_{0.8}$ & 56.2 \\ \cline{2-9} 
& Relative Gain & 0.155 & 0.171 & 0.159 & 0.194 & 0.165 & $\mathbf{0.203_2}$ & 0.174 \\
\hline \hline
\multirow{7}{*}{\textbf{t-score}} & Utility F1 & $76.5_{1.7}$ & $75.8_{5.0}$ & $75.9_{1.8}$ & $76.4_{3.6}$ & $76.6_{1.8}$ & $89.5_{0.7}$ & 78.5 \\
& CS & 0.366 & 0.366 & 0.366 & 0.421 & 0.366 & 0.425 & 0.385 \\
& Perplexity & 1479 & 1487 & 1467 & 1404 & 1474 & 1061 & 1395 \\ \cline{2-9} 
& PI Masking & 0.85 & 0.88 & 0.86 & 2.90 & 1.05 & 8.27 & 2.47 \\
& Static F1 & $44.7_{4.2}$ & $41.3_{2.1}$ & $40.2_{1.0}$ & $44.0_{1.8}$ & $42.5_{2.7}$ & $49.6_{3.4}$ & 43.7 \\
& Adaptive F1 & $54.6_{0.6}$ & $55.7_{1.2}$ & $55.5_{2.9}$ & $57.0_{0.8}$ & $57.0_{2.4}$ & $59.0_{1.9}$ & 56.5 \\ \cline{2-9} 
& Relative Gain & 0.161 & 0.166 & 0.172 & 0.173 & 0.159 & $\mathbf{0.187_2}$ & 0.170 \\
\hline \hline
\multirow{7}{*}{\textbf{POS}} & Utility F1 & $82.5_{2.2}$ & $85.7_{1.0}$ & $84.9_{1.7}$ & $86.2_{0.9}$ & $85.0_{1.4}$ & $92.6_{0.4}$ & 86.2 \\
& CS & 0.504 & 0.504 & 0.505 & 0.545 & 0.506 & 0.574 & 0.523 \\
& Perplexity & 1356 & 1368 & 1373 & 1267 & 1362 & 992 & 1286 \\ \cline{2-9} 
& PI Masking & 5.81 & 5.87 & 5.87 & 8.13 & 6.01 & 11.36 & 7.17 \\
& Static F1 & $49.0_{1.9}$ & $46.7_{3.6}$ & $49.7_{3.5}$ & $51.5_{3.1}$ & $48.1_{9.1}$ & $55.6_{7.5}$ & 50.1 \\
& Adaptive F1 & $58.8_{1.1}$ & $58.4_{0.7}$ & $58.9_{3.0}$ & $57.0_{3.8}$ & $58.5_{1.9}$ & $63.1_{1.2}$ & 59.1 \\ \cline{2-9}
& Relative Gain & $\mathbf{0.205_1}$ & $\mathbf{0.233_{1,2}}$ & $\mathbf{0.215_1}$ & 0.232 & $\mathbf{0.223_1}$ & 0.224 & \textbf{0.222} \\
\hline \hline
\multirow{7}{*}{\textbf{WordNet}} & Utility F1 & $78.7_{1.4}$ & $77.6_{2.6}$ & $74.8_{1.9}$ & $79.2_{4.3}$ & $77.8_{1.7}$ & $90.2_{1.2}$ & 79.7 \\
& CS & 0.336 & 0.335 & 0.336 & 0.413 & 0.335 & 0.418 & 0.362 \\
& Perplexity & 2063 & 2065 & 2078 & 1850 & 2034 & 1375 & 1911 \\ \cline{2-9} 
& PI Masking & 1.44 & 1.49 & 1.28 & 4.25 & 1.48 & 9.57 & 3.25 \\
& Static F1 & $38.7_{2.8}$ & $42.1_{7.5}$ & $37.5_{0.4}$ & $35.3_{5.1}$ & $37.6_{0.3}$ & $40.8_{1.8}$ & 38.7 \\
& Adaptive F1 & $57.6_{0.3}$ & $57.1_{1.2}$ & $57.0_{1.4}$ & $49.8_{12.8}$ & $57.2_{0.1}$ & $54.0_{8.4}$ & 55.4 \\ \cline{2-9} 
& Relative Gain & 0.166 & 0.148 & 0.154 & $\underline{\mathbf{0.242_{1,2}}}$ & 0.167 & $\mathbf{0.239_2}$ & 0.186 \\
\hline
\multirow{7}{*}{\textbf{Average}} & \textbf{Utility F1} & 77.9 & 76.3 & 77.4 & 78.5 & 78.0 & 89.9 & -- \\
 & \textbf{CS} & 0.391 & 0.390 & 0.391 & 0.448 & 0.391 & 0.457 & -- \\
 & \textbf{Perplexity} & 1549 & 1557 & 1552 & 1441 & 1537 & 1107 & -- \\ \cline{2-9} 
 & \textbf{PI Masking} & 1.93 & 1.98 & 1.94 & 4.22 & 2.10 & 9.15 & -- \\
 & \textbf{Static F1} & 44.2 & 42.3 & 42.3 & 43.9 & 43.6 & 46.9 & -- \\
 & \textbf{Adaptive F1} & 56.7 & 56.6 & 56.7 & 55.3 & 56.4 & 58.5 & -- \\ \cline{2-9} 
 & \textbf{Relative Gain} & 0.169 & 0.169 & 0.175 & 0.199 & 0.173 & \textbf{0.215} & -- \\
\hline
\end{tabular}%
}
\caption{Results for the Trustpilot dataset at $\varepsilon=5.2$. Each combination of decomposition and distribution strategy yields scores for Utility F1 ($\uparrow$), Cosine Similarity (\textit{CS}, $\uparrow$), Perplexity ($\downarrow$), PI Masking ($\downarrow$), Static F1 ($\downarrow$), Adaptive F1 ($\downarrow$), and Relative Gain ($\uparrow$). Values derived from an average of three runs are displayed as mean$_{\text{std}}$. F1 values are in \%. The rightmost column and bottom rows calculate the average results over decomposition and distribution methods, respectively. Best average Relative Gain values are \textbf{bolded}. For individual Relative Gain values, only the best are bolded: $\mathbf{x}_1$ for best per decomposition, $\mathbf{x}_2$ for best per budget distribution, and $\mathbf{x}_{1,2}$ if both. The global best Relative Gain (across all combinations) is \underline{underlined}.}
\label{tab:results_trustpilot_5_2}
\end{table*}

\begin{table*}[htbp]
\centering
\resizebox{\textwidth}{!}{
\begin{tabular}{l|l|cccccc|c}
\hline
\multicolumn{2}{c|}{\multirow{2}{*}{\begin{tabular}{c}\textbf{Trustpilot} \\ $\varepsilon=52$\end{tabular}}} & \multicolumn{6}{c|}{\textbf{Budget Distribution}} & \multirow{3}{*}{\begin{tabular}{c}\textbf{Average}\end{tabular}} \\
\cline{3-8}
\multicolumn{2}{c|}{}  & \textbf{Baseline} & \textbf{Attention} & \textbf{Gradients} & \textbf{IC} & \textbf{KeyBERT} & \textbf{YAKE} &  \\
\cline{1-8}
\textbf{Decomposition} & \textbf{Baseline} & \multicolumn{6}{c|}{Utility: $98.2_{0.4}$ / Adversary: $72.1_{1.0}$} &  \\
\hline
\multirow{7}{*}{\textbf{PMI}} & Utility F1 &$82.0_{0.8}$ & $80.6_{0.6}$ & $84.0_{1.6}$ & $79.1_{0.5}$ & $82.4_{2.6}$ & $93.8_{1.2}$ & 83.6 \\
& CS & 0.525 & 0.521 & 0.521 & 0.577 & 0.492 & 0.614 & 0.542 \\
& Perplexity & 801 & 794 & 787 & 706 & 770 & 515 & 729 \\ \cline{2-9} 
&  PI Masking & 6.19 & 7.45 & 7.40 & 7.80 & 9.02 & 14.11 & 8.66 \\
& Static F1 & $49.2_{2.0}$ & $48.0_{1.8}$ & $48.8_{2.0}$ & $52.5_{0.8}$ & $49.5_{1.9}$ & $56.3_{2.9}$ & 50.7 \\
& Adaptive F1 & $58.9_{0.8}$ & $59.6_{0.9}$ & $59.6_{2.2}$ & $58.9_{2.9}$ & $58.6_{0.8}$ & $60.5_{4.7}$ & 59.4 \\ \cline{2-9} 
& Relative Gain & 0.211 & 0.199 & 0.213 & 0.202 & 0.184 & $\mathbf{0.247_2}$ & 0.209 \\
\hline \hline
\multirow{7}{*}{\textbf{LLR}} & Utility F1 & $83.5_{2.0}$ & $81.3_{1.6}$ & $82.9_{1.8}$ & $81.7_{3.9}$ & $83.0_{3.6}$ & $94.4_{0.3}$ & 84.5 \\
& CS & 0.529 & 0.527 & 0.527 & 0.583 & 0.496 & 0.616 & 0.546 \\
& Perplexity & 765 & 763 & 753 & 681 & 746 & 495 & 701 \\ \cline{2-9} 
&  PI Masking & 6.00 & 6.81 & 6.92 & 8.49 & 8.46 & 13.74 & 8.40 \\
& Static F1 & $48.2_{1.3}$ & $51.3_{5.1}$ & $50.5_{1.4}$ & $54.9_{0.7}$ & $44.4_{6.4}$ & $56.4_{4.0}$ & 51.0 \\
& Adaptive F1 & $55.3_{5.0}$ & $53.6_{8.1}$ & $56.0_{3.4}$ & $58.2_{3.4}$ & $54.3_{2.8}$ & $60.5_{4.8}$ & 56.3 \\ \cline{2-9} 
& Relative Gain & 0.240 & 0.219 & 0.220 & 0.209 & 0.231 & $\mathbf{0.252_2}$ & 0.228 \\
\hline \hline
\multirow{7}{*}{\textbf{t-score}} & Utility F1 & $82.3_{2.1}$ & $84.9_{2.2}$ & $84.3_{1.7}$ & $84.1_{0.5}$ & $82.6_{0.6}$ & $94.3_{1.3}$ & 85.4 \\ 
& CS & 0.525 & 0.522 & 0.522 & 0.580 & 0.492 & 0.616 & 0.543 \\
& Perplexity & 763 & 758 & 754 & 683 & 730 & 493 & 697 \\ \cline{2-9} 
& PI Masking & 6.66 & 7.95 & 8.05 & 9.36 & 9.92 & 14.89 & 9.47 \\
& Static F1 & $47.3_{0.8}$ & $48.0_{0.6}$ & $46.2_{8.3}$ & $52.0_{1.9}$ & $48.0_{2.6}$ & $59.3_{1.3}$ & 50.1 \\
& Adaptive F1 & $50.1_{12.7}$ & $54.7_{6.0}$ & $60.4_{0.4}$ & $60.2_{1.6}$ & $57.6_{1.3}$ & $60.7_{3.5}$ & 57.3 \\ \cline{2-9} 
& Relative Gain & $\mathbf{0.254_{1,2}}$ & 0.239 & 0.220 & 0.219 & 0.192 & 0.234 & 0.226 \\
\hline \hline
\multirow{7}{*}{\textbf{POS}} & Utility F1 & $88.6_{1.7}$ & $90.6_{1.2}$ & $86.4_{0.4}$ & $88.1_{0.5}$ & $85.6_{1.8}$ & $95.7_{1.0}$ & 89.2 \\
& CS & 0.618 & 0.618 & 0.618 & 0.655 & 0.603 & 0.702 & 0.636 \\
& Perplexity & 898 & 882 & 869 & 792 & 841 & 534 & 803 \\ \cline{2-9} 
& PI Masking & 10.42 & 10.98 & 10.83 & 12.47 & 12.40 & 16.13 & 12.20 \\
& Static F1 & $53.6_{2.6}$ & $54.3_{2.7}$ & $54.0_{2.8}$ & $54.8_{2.9}$ & $47.0_{8.8}$ & $57.7_{2.3}$ & 53.6 \\
& Adaptive F1 & $63.7_{0.3}$ & $58.1_{3.5}$ & $62.5_{0.7}$ & $61.6_{0.5}$ & $62.6_{0.7}$ & $65.8_{0.7}$ & 62.4 \\ \cline{2-9}
& Relative Gain & 0.236 & $\mathbf{0.264_1}$ & 0.227 & 0.248 & $\mathbf{0.237_1}$ & $\underline{\mathbf{0.265_{1,2}}}$ & \textbf{0.246} \\
\hline \hline
\multirow{7}{*}{\textbf{WordNet}} & Utility F1 & $85.7_{1.3}$ & $83.8_{1.7}$ & $85.1_{1.5}$ & $88.8_{1.3}$ & $83.4_{0.9}$ & $94.8_{1.1}$ & 86.9 \\
& CS & 0.529 & 0.522 & 0.523 & 0.595 & 0.482 & 0.642 & 0.549 \\
& Perplexity & 939 & 942 & 932 & 793 & 931 & 476 & 836 \\ \cline{2-9} 
& PI Masking & 4.09 & 4.32 & 4.29 & 7.87 & 5.67 & 17.46 & 7.29 \\
& Static F1 & $46.7_{1.3}$ & $48.3_{2.4}$ & $48.0_{1.9}$ & $53.9_{4.9}$ & $43.1_{1.6}$ & $56.0_{1.3}$ & 49.3 \\
& Adaptive F1 & $61.0_{0.7}$ & $58.0_{2.2}$ & $60.3_{1.3}$ & $59.4_{1.9}$ & $60.8_{1.0}$ & $63.6_{3.6}$ & 60.5 \\\cline{2-9} 
& Relative Gain & 0.242 & 0.233 & $\mathbf{0.232_1}$ & $\mathbf{0.252_{1,2}}$ & 0.216 & 0.241 & 0.236 \\
\hline
\multirow{7}{*}{\textbf{Average}} & \textbf{Utility F1} & 84.4 & 84.2 & 84.5 & 84.4 & 83.4 & 94.6 & -- \\
 & \textbf{CS} & 0.545 & 0.542 & 0.542 & 0.598 & 0.513 & 0.638 & -- \\
 & \textbf{Perplexity} & 833 & 828 & 819 & 731 & 803 & 503 & -- \\ \cline{2-9} 
 & \textbf{PI Masking} & 6.67 & 7.50 & 7.50 & 9.20 & 9.09 & 15.27 & -- \\
 & \textbf{Static F1} & 49.0 & 50.0 & 49.5 & 53.6 & 46.4 & 57.2 & -- \\
 & \textbf{Adaptive F1} & 57.8 & 56.8 & 59.8 & 59.7 & 58.8 & 62.2 & -- \\ \cline{2-9} 
 & \textbf{Relative Gain} & 0.236 & 0.231 & 0.222 & 0.226 & 0.212 & \textbf{0.248} & -- \\
\hline
\end{tabular}
}
\caption{Results for the Trustpilot dataset at $\varepsilon=52$. Each combination of decomposition and distribution strategy yields scores for Utility F1 ($\uparrow$), Cosine Similarity (\textit{CS}, $\uparrow$), Perplexity ($\downarrow$), PI Masking ($\downarrow$), Static F1 ($\downarrow$), Adaptive F1 ($\downarrow$), and Relative Gain ($\uparrow$). Values derived from an average of three runs are displayed as mean$_{\text{std}}$. F1 values are in \%. The rightmost column and bottom rows calculate the average results over decomposition and distribution methods, respectively. Best average Relative Gain values are \textbf{bolded}. For individual Relative Gain values, only the best are bolded: $\mathbf{x}_1$ for best per decomposition, $\mathbf{x}_2$ for best per budget distribution, and $\mathbf{x}_{1,2}$ if both. The global best Relative Gain (across all combinations) is \underline{underlined}.}
\label{tab:results_trustpilot_52}
\end{table*}

\begin{table*}[htbp]
\centering
\resizebox{\textwidth}{!}{
\begin{tabular}{l|l|cccccc|c}
\hline
\multicolumn{2}{c|}{\multirow{2}{*}{\begin{tabular}{c}\textbf{Trustpilot} \\ $\varepsilon=260$\end{tabular}}} & \multicolumn{6}{c|}{\textbf{Budget Distribution}} & \multirow{3}{*}{\begin{tabular}{c}\textbf{Average}\end{tabular}} \\
\cline{3-8}
\multicolumn{2}{c|}{}  & \textbf{Baseline} & \textbf{Attention} & \textbf{Gradients} & \textbf{IC} & \textbf{KeyBERT} & \textbf{YAKE} &  \\
\cline{1-8}
\textbf{Decomposition} & \textbf{Baseline} & \multicolumn{6}{c|}{Utility: $98.2_{0.4}$ / Adversary: $72.1_{1.0}$} &  \\
\hline
\multirow{7}{*}{\textbf{PMI}} & Utility F1 & $96.3_{0.4}$ & $96.5_{0.4}$ & $96.1_{0.7}$ & $95.7_{1.3}$ & $97.1_{0.0}$ & $98.4_{0.4}$ & 96.7 \\
& CS & 0.855 & 0.841 & 0.835 & 0.864 & 0.759 & 0.891 & 0.841 \\
& Perplexity & 220 & 230 & 230 & 212 & 245 & 171 & 218 \\ \cline{2-9} 
&  PI Masking & 29.72 & 30.62 & 30.14 & 30.07 & 31.65 & 35.48 & 31.28 \\
& Static F1 & $64.7_{3.7}$ & $67.8_{0.3}$ & $66.4_{1.9}$ & $66.0_{0.6}$ & $63.8_{3.0}$ & $57.5_{17.8}$ & 64.4 \\
& Adaptive F1 & $61.4_{9.3}$ & $69.3_{1.0}$ & $67.1_{3.9}$ & $69.5_{0.9}$ & $66.8_{1.4}$ & $69.1_{1.4}$ & 67.2 \\ \cline{2-9} 
& Relative Gain & 0.280 & 0.225 & 0.236 & 0.241 & 0.208 & $\mathbf{0.282_{1,2}}$ & 0.245 \\
\hline \hline
\multirow{7}{*}{\textbf{LLR}} & Utility F1 & $95.5_{0.3}$ & $96.1_{0.9}$ & $95.6_{0.3}$ & $96.1_{0.3}$ & $96.7_{0.5}$ & $98.0_{0.5}$ & 96.3 \\
& CS & 0.849 & 0.835 & 0.828 & 0.860 & 0.755 & 0.888 & 0.836 \\
& Perplexity & 226 & 232 & 235 & 217 & 246 & 171 & 221 \\ \cline{2-9} 
&  PI Masking & 28.23 & 29.22 & 28.91 & 28.94 & 30.07 & 34.30 & 29.94 \\
& Static F1 & $62.6_{8.4}$ & $65.2_{2.1}$ & $64.7_{1.0}$ & $66.5_{1.3}$ & $64.9_{1.5}$ & $68.0_{0.6}$ & 65.3 \\
& Adaptive F1 & $68.2_{1.9}$ & $67.4_{1.9}$ & $64.0_{2.5}$ & $70.1_{1.1}$ & $65.4_{2.7}$ & $68.5_{4.1}$ & 67.3 \\ \cline{2-9} 
& Relative Gain & $\mathbf{0.260_2}$ & 0.243 & $\mathbf{0.255_1}$ & 0.241 & 0.212 & 0.243 & 0.242 \\
\hline \hline
\multirow{7}{*}{\textbf{t-score}} & Utility F1 & $96.3_{1.1}$ & $95.7_{1.1}$ & $94.9_{0.4}$ & $97.5_{0.3}$ & $96.6_{0.9}$ & $98.4_{0.6}$ & 96.6 \\ 
& CS & 0.863 & 0.848 & 0.840 & 0.872 & 0.763 & 0.899 & 0.848 \\
& Perplexity & 215 & 220 & 226 & 207 & 241 & 166 & 212 \\ \cline{2-9} 
& PI Masking & 31.90 & 32.28 & 32.13 & 32.05 & 33.36 & 36.87 & 33.10 \\
& Static F1 & $68.3_{1.3}$ & $67.0_{0.6}$ & $67.6_{1.2}$ & $68.1_{1.4}$ & $66.9_{0.7}$ & $68.5_{1.3}$ & 67.7 \\
& Adaptive F1 & $67.4_{1.8}$ & $56.9_{18.8}$ & $68.1_{0.8}$ & $69.1_{1.7}$ & $67.4_{2.3}$ & $64.0_{9.1}$ & 65.5 \\ \cline{2-9} 
& Relative Gain & 0.235 & $\mathbf{0.271_{1,2}}$ & 0.216 & 0.239 & 0.186 & 0.257 & 0.234 \\
\hline \hline
\multirow{7}{*}{\textbf{POS}} & Utility F1 & $97.3_{0.7}$ & $96.5_{0.3}$ & $94.7_{0.4}$ & $95.9_{0.9}$ & $96.6_{0.5}$ & $97.0_{1.0}$ & 96.3 \\
& CS & 0.829 & 0.822 & 0.821 & 0.837 & 0.784 & 0.868 & 0.827 \\
& Perplexity & 352 & 349 & 354 & 336 & 356 & 269 & 336 \\ \cline{2-9} 
& PI Masking & 26.84 & 27.20 & 27.70 & 28.11 & 29.09 & 32.14 & 28.51 \\
& Static F1 & $64.1_{1.5}$ & $65.0_{2.1}$ & $66.1_{0.8}$ & $67.5_{0.4}$ & $63.3_{2.6}$ & $66.4_{2.5}$ & 65.4 \\
& Adaptive F1 & $67.5_{1.6}$ & $62.5_{8.5}$ & $68.5_{1.2}$ & $69.0_{0.1}$ & $66.5_{2.5}$ & $70.0_{0.1}$ & 67.3 \\ \cline{2-9}
& Relative Gain & 0.261 & $\mathbf{0.268_2}$ & 0.228 & 0.232 & $\mathbf{0.233_1}$ & 0.237 & 0.243 \\
\hline \hline
\multirow{7}{*}{\textbf{WordNet}} & Utility F1 & $96.7_{0.4}$ & $97.5_{0.4}$ & $95.9_{0.7}$ & $96.1_{0.8}$ & $96.6_{0.2}$ & $97.5_{0.4}$ & 96.7 \\
& CS & 0.833 & 0.820 & 0.810 & 0.849 & 0.735 & 0.881 & 0.821 \\
& Perplexity & 257 & 260 & 267 & 238 & 281 & 169 & 245 \\ \cline{2-9} 
& PI Masking & 22.26 & 23.54 & 24.09 & 25.20 & 25.96 & 37.93 & 26.50 \\
& Static F1 & $65.7_{0.7}$ & $66.1_{1.8}$ & $63.5_{2.3}$ & $63.8_{2.6}$ & $61.3_{1.9}$ & $68.0_{0.8}$ & 64.7 \\
& Adaptive F1 & $63.2_{10.2}$ & $69.0_{1.7}$ & $69.1_{2.0}$ & $68.1_{1.5}$ & $67.7_{1.4}$ & $66.6_{4.1}$ & 67.3 \\\cline{2-9} 
& Relative Gain & $\underline{\mathbf{0.289_{1,2}}}$ & 0.256 & 0.251 & $\mathbf{0.270_1}$ & 0.224 & 0.230 & \textbf{0.253} \\
\hline
\multirow{7}{*}{\textbf{Average}} & \textbf{Utility F1} & 96.4 & 96.5 & 95.4 & 96.2 & 96.7 & 97.9 & -- \\
 & \textbf{CS} & 0.846 & 0.833 & 0.827 & 0.856 & 0.759 & 0.885 & -- \\
 & \textbf{Perplexity} & 254 & 258 & 262 & 242 & 274 & 189 & -- \\ \cline{2-9} 
 & \textbf{PI Masking} & 27.79 & 28.57 & 28.59 & 28.87 & 30.02 & 35.34 & -- \\
 & \textbf{Static F1} & 65.1 & 66.2 & 65.7 & 66.4 & 64.1 & 65.7 & -- \\
 & \textbf{Adaptive F1} & 65.5 & 65.0 & 67.4 & 69.2 & 66.8 & 67.7 & -- \\ \cline{2-9} 
 & \textbf{Relative Gain} & \textbf{0.265} & 0.253 & 0.237 & 0.245 & 0.213 & 0.250 & -- \\
\hline
\end{tabular}
}
\caption{Results for the Trustpilot dataset at $\varepsilon=260$. Each combination of decomposition and distribution strategy yields scores for Utility F1 ($\uparrow$), Cosine Similarity (\textit{CS}, $\uparrow$), Perplexity ($\downarrow$), PI Masking ($\downarrow$), Static F1 ($\downarrow$), Adaptive F1 ($\downarrow$), and Relative Gain ($\uparrow$). Values derived from an average of three runs are displayed as mean$_{\text{std}}$. F1 values are in \%. The rightmost columns and bottom rows calculate the average results over decomposition and distribution methods, respectively. Best average Relative Gain values are \textbf{bolded}. For individual Relative Gain values, only the best are bolded: $\mathbf{x}_1$ for best per decomposition, $\mathbf{x}_2$ for best per budget distribution, and $\mathbf{x}_{1,2}$ if both. The global best Relative Gain (across all combinations) is \underline{underlined}.}
\label{tab:results_trustpilot_260}
\end{table*}

\begin{table*}[htbp]
\centering
\resizebox{\textwidth}{!}{
\begin{tabular}{l|l|cccccc|c}
\hline
\multicolumn{2}{c|}{\multirow{2}{*}{\begin{tabular}{c}\textbf{Yelp} \\ $\varepsilon=18.7$\end{tabular}}} & \multicolumn{6}{c|}{\textbf{Budget Distribution}} & \multirow{3}{*}{\begin{tabular}{c}\textbf{Average}\end{tabular}} \\
\cline{3-8}
\multicolumn{2}{c|}{}  & \textbf{Baseline} & \textbf{Attention} & \textbf{Gradients} & \textbf{IC} & \textbf{KeyBERT} & \textbf{YAKE} &  \\
\cline{1-8}
\textbf{Decomposition} & \textbf{Baseline} & \multicolumn{6}{c|}{Utility: $87.8_{1.9}$ / Adversary: $94.4_{0.2}$} &  \\
\hline
\multirow{7}{*}{\textbf{PMI}} & Utility F1 & $48.1_{0.0}$ & $48.1_{0.0}$ & $48.1_{0.0}$ & $48.1_{0.0}$ & $48.1_{0.0}$ & $76.6_{3.2}$ & 52.8 \\
& CS & 0.348 & 0.349 & 0.349 & 0.380 & 0.348 & 0.590 & 0.394 \\
& Perplexity & 717 & 692 & 693 & 702 & 687 & 322 & 635 \\ \cline{2-9} 
&  PI Masking & 0.37 & 0.48 & 0.57 & 0.81 & 0.56 & 14.22 & 2.83 \\
& Static F1 & $14.8_{2.2}$ & $15.3_{3.1}$ & $14.8_{2.5}$ & $14.9_{2.8}$ & $13.6_{2.6}$ & $55.5_{5.1}$ & 21.5 \\
& Adaptive F1 & $77.6_{1.7}$ & $76.5_{2.3}$ & $77.1_{0.8}$ & $77.7_{1.0}$ & $77.4_{1.2}$ & $88.7_{1.1}$ & 79.2 \\ \cline{2-9} 
& Relative Gain & 0.120 & 0.122 & 0.122 & 0.135 & 0.124 & $\underline{\mathbf{0.174_{1,2}}}$ & 0.133 \\
\hline \hline
\multirow{7}{*}{\textbf{LLR}} & Utility F1 & $48.1_{0.0}$ & $48.1_{0.0}$ & $48.1_{0.0}$ & $48.1_{0.0}$ & $48.1_{0.0}$ & $58.3_{17.7}$ & 49.8 \\
& CS & 0.348 & 0.349 & 0.348 & 0.383 & 0.347 & 0.594 & 0.395 \\
& Perplexity & 693 & 668 & 693 & 674 & 689 & 314 & 622 \\ \cline{2-9} 
&  PI Masking & 0.37 & 0.53 & 0.44 & 0.88 & 0.58 & 14.38 & 2.86 \\
& Static F1 & $12.8_{2.7}$ & $12.4_{2.8}$ & $12.7_{3.3}$ & $13.3_{2.9}$ & $13.5_{3.2}$ & $53.7_{5.8}$ & 19.8 \\
& Adaptive F1 & $75.7_{2.2}$ & $76.2_{2.3}$ & $76.5_{2.0}$ & $77.2_{2.3}$ & $75.6_{4.6}$ & $87.4_{1.5}$ & 78.1 \\ \cline{2-9} 
& Relative Gain & 0.133 & 0.133 & 0.131 & $\mathbf{0.143_2}$ & 0.130 & 0.088 & 0.126 \\
\hline \hline
\multirow{7}{*}{\textbf{t-score}} & Utility F1 & $48.1_{0.0}$ & $48.1_{0.0}$ & $48.1_{0.0}$ & $48.1_{0.0}$ & $48.1_{0.0}$ & $69.1_{18.2}$ & 51.6 \\ 
& CS & 0.362 & 0.363 & 0.362 & 0.396 & 0.362 & 0.591 & 0.406 \\
& Perplexity & 751 & 754 & 720 & 727 & 745 & 312 & 668 \\ \cline{2-9} 
& PI Masking & 0.45 & 0.55 & 0.54 & 0.86 & 0.59 & 14.38 & 2.90 \\
& Static F1 & $12.2_{2.0}$ & $12.6_{2.1}$ & $12.1_{2.8}$ & $13.5_{2.5}$ & $12.7_{2.0}$ & $55.9_{4.2}$ & 19.8 \\
& Adaptive F1 & $76.7_{3.3}$ & $77.1_{1.7}$ & $77.4_{0.7}$ & $77.6_{1.4}$ & $76.6_{2.7}$ & $88.0_{1.7}$ & 78.9 \\ \cline{2-9} 
& Relative Gain & 0.139 & 0.137 & 0.137 & $\mathbf{0.148_{1,2}}$ & 0.137 & 0.135 & 0.139 \\
\hline \hline
\multirow{7}{*}{\textbf{POS}} & Utility F1 & $48.1_{0.0}$ & $48.1_{0.0}$ & $48.1_{0.0}$ & $48.1_{0.0}$ & $48.1_{0.0}$ & $65.2_{15.9}$ & 50.9 \\
& CS & 0.452 & 0.453 & 0.454 & 0.472 & 0.453 & 0.712 & 0.499 \\
& Perplexity & 596 & 594 & 567 & 578 & 589 & 281 & 534 \\ \cline{2-9} 
& PI Masking & 4.18 & 4.34 & 4.30 & 4.56 & 4.33 & 15.59 & 6.22 \\
& Static F1 & $18.2_{3.9}$ & $17.6_{3.5}$ & $17.2_{3.4}$ & $19.0_{3.1}$ & $18.2_{3.4}$ & $59.6_{4.5}$ & 24.9 \\
& Adaptive F1 & $78.6_{2.6}$ & $81.6_{0.4}$ & $80.7_{2.8}$ & $80.5_{0.5}$ & $79.5_{1.1}$ & $89.0_{1.4}$ & 81.6 \\ \cline{2-9}
& Relative Gain & $\mathbf{0.147_1}$ & $\mathbf{0.139_1}$ & $\mathbf{0.144_1}$ & 0.147 & $\mathbf{0.144_1}$ & $\mathbf{0.158_2}$ & \textbf{0.146} \\
\hline \hline
\multirow{7}{*}{\textbf{WordNet}} & Utility F1 & $48.1_{0.0}$ & $48.1_{0.0}$ & $48.1_{0.0}$ & $48.1_{0.0}$ & $48.1_{0.0}$ & $69.6_{18.7}$ & 51.6 \\
& CS & 0.323 & 0.324 & 0.324 & 0.358 & 0.323 & 0.593 & 0.374 \\
& Perplexity & 827 & 825 & 852 & 795 & 811 & 296 & 734 \\ \cline{2-9} 
& PI Masking & 0.55 & 0.72 & 0.72 & 1.07 & 0.81 & 15.70 & 3.26 \\
& Static F1 & $15.1_{4.7}$ & $14.6_{3.2}$ & $14.3_{3.8}$ & $15.4_{3.7}$ & $14.4_{3.4}$ & $62.6_{4.8}$ & 22.7 \\
& Adaptive F1 & $78.8_{2.3}$ & $80.3_{1.6}$ & $80.7_{0.5}$ & $79.0_{0.4}$ & $78.7_{1.3}$ & $90.4_{0.2}$ & 81.3 \\\cline{2-9} 
& Relative Gain & 0.101 & 0.098 & 0.097 & $\mathbf{0.116_2}$ & 0.103 & 0.102 & 0.103 \\
\hline
\multirow{7}{*}{\textbf{Average}} & \textbf{Utility F1} & 48.1 & 48.1 & 48.1 & 48.1 & 48.1 & 67.8 & -- \\
 & \textbf{CS} & 0.367 & 0.368 & 0.367 & 0.398 & 0.367 & 0.616 & -- \\
 & \textbf{Perplexity} & 717 & 706 & 705 & 695 & 704 & 305 & -- \\ \cline{2-9} 
 & \textbf{PI Masking} & 1.18 & 1.32 & 1.31 & 1.64 & 1.37 & 14.85 & -- \\
 & \textbf{Static F1} & 14.6 & 14.5 & 14.2 & 15.2 & 14.5 & 57.5 & -- \\
 & \textbf{Adaptive F1} & 77.5 & 78.3 & 78.5 & 78.4 & 77.6 & 88.7 & -- \\ \cline{2-9} 
 & \textbf{Relative Gain} & 0.128 & 0.126 & 0.126 & \textbf{0.138} & 0.128 & 0.131 & -- \\
\hline
\end{tabular}
}
\caption{Results for the Yelp dataset at $\varepsilon=18.7$. Each combination of decomposition and distribution strategy yields scores for Utility F1 ($\uparrow$), Cosine Similarity (\textit{CS}, $\uparrow$), Perplexity ($\downarrow$), PI Masking ($\downarrow$), Static F1 ($\downarrow$), Adaptive F1 ($\downarrow$), and Relative Gain ($\uparrow$). Values derived from an average of three runs are displayed as mean$_{\text{std}}$. F1 values are in \%. The rightmost column and bottom rows calculate the average results over decomposition and distribution methods, respectively. Best average Relative Gain values are \textbf{bolded}. For individual Relative Gain values, only the best are bolded: $\mathbf{x}_1$ for best per decomposition, $\mathbf{x}_2$ for best per budget distribution, and $\mathbf{x}_{1,2}$ if both. The global best Relative Gain (across all combinations) is \underline{underlined}.}
\label{tab:results_trustpilot_18_7}
\end{table*}

\begin{table*}[htbp]
\centering
\resizebox{\textwidth}{!}{
\begin{tabular}{l|l|cccccc|c}
\hline
\multicolumn{2}{c|}{\multirow{2}{*}{\begin{tabular}{c}\textbf{Yelp} \\ $\varepsilon=187$\end{tabular}}} & \multicolumn{6}{c|}{\textbf{Budget Distribution}} & \multirow{3}{*}{\begin{tabular}{c}\textbf{Average}\end{tabular}} \\
\cline{3-8}
\multicolumn{2}{c|}{}  & \textbf{Baseline} & \textbf{Attention} & \textbf{Gradients} & \textbf{IC} & \textbf{KeyBERT} & \textbf{YAKE} &  \\
\cline{1-8}
\textbf{Decomposition} & \textbf{Baseline} & \multicolumn{6}{c|}{Utility: $87.8_{1.9}$ / Adversary: $94.4_{0.2}$} &  \\
\hline
\multirow{7}{*}{\textbf{PMI}} & Utility F1 & $48.1_{0.0}$ & $48.1_{0.0}$ & $48.1_{0.0}$ & $48.1_{0.0}$ & $51.1_{5.3}$ & $85.3_{2.6}$ & 54.8  \\
& CS & 0.557 & 0.549 & 0.550 & 0.578 & 0.505 & 0.829 & 0.595 \\
& Perplexity & 484 & 486 & 455 & 463 & 480 & 144 & 419 \\ \cline{2-9} 
&  PI Masking & 6.17 & 6.61 & 6.45 & 6.45 & 6.83 & 22.66 & 9.19 \\
& Static F1 & $29.7_{4.5}$ & $28.5_{5.9}$ & $29.0_{4.4}$ & $30.8_{5.5}$ & $29.8_{4.9}$ & $81.9_{1.8}$ & 38.3 \\
& Adaptive F1 & $80.9_{0.8}$ & $79.5_{2.9}$ & $80.8_{1.6}$ & $81.0_{2.9}$ & $79.4_{0.5}$ & $90.6_{1.5}$ & 82.0 \\ \cline{2-9} 
& Relative Gain & 0.148 & 0.152 & 0.146 & 0.154 & 0.139 & $\mathbf{0.220_2}$ & 0.160 \\
\hline \hline
\multirow{7}{*}{\textbf{LLR}} & Utility F1 & $48.1_{0.0}$ & $48.1_{0.0}$ & $48.1_{0.0}$ & $51.9_{6.7}$ & $48.1_{0.0}$ & $57.9_{12.8}$ & 50.3 \\
& CS & 0.549 & 0.544 & 0.543 & 0.572 & 0.499 & 0.601 & 0.551 \\
& Perplexity & 491 & 492 & 491 & 471 & 485 & 352 & 464 \\ \cline{2-9} 
&  PI Masking & 5.31 & 5.98 & 5.84 & 6.06 & 6.37 & 7.92 & 6.25 \\
& Static F1 & $28.2_{5.4}$ & $28.5_{4.5}$ & $29.5_{5.0}$ & $29.8_{5.3}$ & $29.6_{5.2}$ & $39.0_{5.7}$ & 30.8 \\
& Adaptive F1 & $81.4_{1.1}$ & $80.8_{2.4}$ & $79.6_{1.5}$ & $80.0_{3.0}$ & $80.2_{3.0}$ & $83.6_{1.1}$ & 80.9 \\ \cline{2-9} 
& Relative Gain & 0.150 & 0.146 & 0.147 & $\mathbf{0.180_2}$ & 0.119 & 0.176 & 0.153 \\
\hline \hline
\multirow{7}{*}{\textbf{t-score}} & Utility F1 & $48.1_{0.0}$ & $48.1_{0.0}$ & $48.1_{0.0}$ & $48.1_{0.0}$ & $51.0_{3.1}$ & $83.2_{2.1}$ & 54.4 \\ 
& CS & 0.558 & 0.552 & 0.551 & 0.580 & 0.506 & 0.832 & 0.596 \\
& Perplexity & 479 & 480 & 480 & 459 & 474 & 142 & 419 \\ \cline{2-9} 
& PI Masking & 6.42 & 7.02 & 6.60 & 6.82 & 7.35 & 23.06 & 9.54 \\
& Static F1 & $29.0_{4.2}$ & $28.9_{5.2}$ & $28.4_{4.5}$ & $30.8_{5.4}$ & $30.5_{5.7}$ & $82.5_{1.2}$ & 38.3 \\
& Adaptive F1 & $82.0_{0.9}$ & $80.9_{2.0}$ & $82.2_{2.6}$ & $80.9_{1.5}$ & $81.8_{2.2}$ & $90.1_{1.2}$ & 83.0 \\ \cline{2-9} 
& Relative Gain & 0.147 & 0.145 & 0.143 & 0.154 & 0.127 & $\mathbf{0.209_2}$ & 0.154 \\
\hline \hline
\multirow{7}{*}{\textbf{POS}} & Utility F1 & $48.1_{0.0}$ & $53.1_{8.7}$ & $48.1_{0.0}$ & $55.2_{8.4}$ & $59.7_{12.3}$ & $84.8_{0.8}$ & 58.1 \\
& CS & 0.646 & 0.644 & 0.645 & 0.662 & 0.620 & 0.860 & 0.680 \\
& Perplexity & 504 & 504 & 500 & 484 & 489 & 159 & 440 \\ \cline{2-9} 
& PI Masking & 8.06 & 8.72 & 8.35 & 8.77 & 9.05 & 21.36 & 10.72 \\
& Static F1 & $32.7_{7.2}$ & $33.3_{7.7}$ & $32.3_{5.8}$ & $33.7_{7.6}$ & $33.3_{6.6}$ & $78.2_{1.8}$ & 40.6 \\
& Adaptive F1 & $80.0_{1.9}$ & $82.3_{0.5}$ & $83.5_{1.2}$ & $84.6_{1.7}$ & $82.3_{1.5}$ & $90.3_{1.3}$ & 83.8 \\ \cline{2-9}
& Relative Gain & $\mathbf{0.182_1}$ & $\mathbf{0.195_1}$ & $\mathbf{0.169_1}$ & $\mathbf{0.207_1}$ & $\mathbf{0.217_1}$ & $\underline{\mathbf{0.252_{1,2}}}$ & \textbf{0.204} \\
\hline \hline
\multirow{7}{*}{\textbf{WordNet}} & Utility F1 & $56.3_{14.3}$ & $48.1_{0.0}$ & $48.1_{0.0}$ & $57.7_{16.8}$ & $53.5_{7.2}$ & $86.5_{0.5}$ & 58.4 \\
& CS & 0.568 & 0.558 & 0.560 & 0.588 & 0.503 & 0.844 & 0.604 \\
& Perplexity & 521 & 523 & 494 & 496 & 520 & 130 & 447 \\ \cline{2-9} 
& PI Masking & 4.28 & 4.63 & 4.54 & 4.79 & 4.96 & 23.04 & 7.71 \\
& Static F1 & $34.5_{5.1}$ & $32.7_{6.6}$ & $34.4_{5.9}$ & $35.9_{5.5}$ & $34.3_{5.3}$ & $85.3_{0.8}$ & 42.9 \\
& Adaptive F1 & $82.6_{2.1}$ & $82.9_{1.7}$ & $84.6_{1.9}$ & $83.6_{1.1}$ & $82.3_{2.5}$ & $91.2_{0.5}$ & 84.6 \\\cline{2-9} 
& Relative Gain & $\mathbf{0.182_1}$ & 0.137 & 0.126 & 0.190 & 0.132 & $\mathbf{0.219_2}$ & 0.164 \\
\hline
\multirow{7}{*}{\textbf{Average}} & \textbf{Utility F1} & 49.7 & 49.1 & 48.1 & 52.2 & 52.7 & 79.5 & -- \\
 & \textbf{CS} & 0.576 & 0.569 & 0.570 & 0.596 & 0.527 & 0.793 & -- \\
 & \textbf{Perplexity} & 495 & 497 & 484 & 475 & 490 & 185 & -- \\ \cline{2-9} 
 & \textbf{PI Masking} & 6.05 & 6.59 & 6.36 & 6.58 & 6.91 & 19.61 & -- \\
 & \textbf{Static F1} & 30.8 & 30.4 & 30.7 & 32.2 & 31.5 & 73.4 & --\\
 & \textbf{Adaptive F1} & 81.4 & 81.3 & 82.2 & 82.0 & 81.2 & 89.2 & -- \\ \cline{2-9} 
 & \textbf{Relative Gain} & 0.162 & 0.155 & 0.147 & 0.177 & 0.147 & \textbf{0.215} & -- \\
\hline
\end{tabular}
}
\caption{Results for the Yelp dataset at $\varepsilon=187$. Each combination of decomposition and distribution strategy yields scores for Utility F1 ($\uparrow$), Cosine Similarity (\textit{CS}, $\uparrow$), Perplexity ($\downarrow$), PI Masking ($\downarrow$), Static F1 ($\downarrow$), Adaptive F1 ($\downarrow$), and Relative Gain ($\uparrow$). Values derived from an average of three runs are displayed as mean$_{\text{std}}$. F1 values are in \%. The rightmost column and bottom rows calculate the average results over decomposition and distribution methods, respectively. Best average Relative Gain values are \textbf{bolded}. For individual Relative Gain values, only the best are bolded: $\mathbf{x}_1$ for best per decomposition, $\mathbf{x}_2$ for best per budget distribution, and $\mathbf{x}_{1,2}$ if both. The global best Relative Gain (across all combinations) is \underline{underlined}.}
\label{tab:results_trustpilot_187}
\end{table*}

\begin{table*}[htbp]
\centering
\resizebox{\textwidth}{!}{
\begin{tabular}{l|l|cccccc|c}
\hline
\multicolumn{2}{c|}{\multirow{2}{*}{\begin{tabular}{c}\textbf{Yelp} \\ $\varepsilon=935$\end{tabular}}} & \multicolumn{6}{c|}{\textbf{Budget Distribution}} & \multirow{3}{*}{\begin{tabular}{c}\textbf{Average}\end{tabular}} \\
\cline{3-8}
\multicolumn{2}{c|}{}  & \textbf{Baseline} & \textbf{Attention} & \textbf{Gradients} & \textbf{IC} & \textbf{KeyBERT} & \textbf{YAKE} &  \\
\cline{1-8}
\textbf{Decomposition} & \textbf{Baseline} & \multicolumn{6}{c|}{Utility: $87.8_{1.9}$ / Adversary: $94.4_{0.2}$} &  \\
\hline
\multirow{7}{*}{\textbf{PMI}} & Utility F1 & $84.9_{1.7}$ & $84.0_{1.5}$ & $83.0_{2.0}$ & $85.1_{2.5}$ & $85.1_{2.1}$ & $86.7_{1.3}$ & 84.8 \\
& CS & 0.887 & 0.874 & 0.873 & 0.883 & 0.796 & 0.962 & 0.879 \\
& Perplexity & 153 & 156 & 156 & 153 & 171 & 99 & 148 \\ \cline{2-9} 
&  PI Masking & 21.10 & 21.30 & 20.83 & 21.36 & 21.85 & 28.31 & 22.46 \\
& Static F1 & $74.8_{1.8}$ & $72.1_{2.5}$ & $73.5_{2.5}$ & $74.9_{2.4}$ & $72.0_{1.9}$ & $88.0_{1.8}$ & $75.9_{0.0}$ \\
& Adaptive F1 & $88.5_{1.6}$ & $89.9_{1.4}$ & $88.2_{0.9}$ & $88.6_{1.6}$ & $90.1_{0.6}$ & $91.2_{0.8}$ & 89.4 \\ \cline{2-9} 
& Relative Gain & $\mathbf{0.286_2}$ & 0.278 & 0.275 & 0.283 & 0.240 & 0.256 & 0.270 \\
\hline \hline
\multirow{7}{*}{\textbf{LLR}} & Utility F1 & $86.0_{0.3}$ & $83.9_{1.6}$ & $86.1_{1.3}$ & $84.5_{0.3}$ & $84.5_{1.5}$ & $87.9_{0.8}$ & 85.5 \\
& CS & 0.881 & 0.868 & 0.868 & 0.878 & 0.788 & 0.962 & 0.874 \\
& Perplexity & 157 & 161 & 161 & 158 & 176 & 99 & 152 \\ \cline{2-9} 
&  PI Masking & 20.34 & 20.45 & 20.20 & 20.64 & 20.78 & 28.35 & 21.79 \\
& Static F1 & $74.1_{1.9}$ & $70.5_{2.5}$ & $72.1_{2.5}$ & $74.3_{2.3}$ & $70.8_{3.1}$ & $87.2_{1.3}$ & 74.8 \\
& Adaptive F1 & $89.9_{1.9}$ & $88.8_{2.1}$ & $90.0_{1.1}$ & $88.7_{1.4}$ & $87.9_{1.1}$ & $91.8_{1.2}$ & 89.5 \\ \cline{2-9} 
& Relative Gain & $\mathbf{0.289_2}$ & 0.286 & $\mathbf{0.289_2}$ & 0.282 & 0.248 & $\mathbf{0.263_1}$ & 0.276 \\
\hline \hline
\multirow{7}{*}{\textbf{t-score}} & Utility F1 & $83.1_{4.3}$ & $82.5_{1.8}$ & $81.9_{2.3}$ & $83.4_{1.8}$ & $81.2_{1.7}$ & $84.3_{1.8}$ & 82.7 \\ 
& CS & 0.891 & 0.877 & 0.878 & 0.887 & 0.796 & 0.966 & 0.882 \\
& Perplexity & 151 & 154 & 154 & 152 & 169 & 98 & 146 \\ \cline{2-9} 
& PI Masking & 21.74 & 21.94 & 21.58 & 21.86 & 22.04 & 28.52 & 22.95 \\
& Static F1 & $76.0_{1.6}$ & $72.9_{2.3}$ & $73.9_{1.7}$ & $76.2_{2.1}$ & $72.4_{2.4}$ & $87.6_{1.8}$ & 76.5 \\
& Adaptive F1 & $89.2_{1.0}$ & $87.7_{1.1}$ & $89.0_{1.8}$ & $88.8_{1.0}$ & $89.3_{1.2}$ & $91.5_{0.9}$ & 89.2 \\ \cline{2-9} 
& Relative Gain & 0.270 & $\mathbf{0.275_2}$ & 0.265 & 0.269 & 0.220 & 0.245 & 0.257 \\
\hline \hline
\multirow{7}{*}{\textbf{POS}} & Utility F1 & $83.1_{2.3}$ & $83.4_{1.1}$ & $80.2_{1.9}$ & $80.9_{1.8}$ & $81.1_{1.6}$ & $84.5_{1.7}$ & 82.2 \\
& CS & 0.869 & 0.863 & 0.863 & 0.869 & 0.823 & 0.941 & 0.871 \\
& Perplexity & 198 & 199 & 199 & 197 & 207 & 122 & 187 \\ \cline{2-9} 
& PI Masking & 18.73 & 19.01 & 18.89 & 18.65 & 19.04 & 25.66 & 20.00 \\
& Static F1 & $64.5_{3.7}$ & $63.4_{3.5}$ & $62.8_{4.2}$ & $64.9_{4.2}$ & $63.1_{4.3}$ & $84.1_{2.0}$ & 67.1 \\
& Adaptive F1 & $87.8_{1.1}$ & $87.0_{1.0}$ & $88.6_{0.5}$ & $89.4_{1.0}$ & $88.2_{0.5}$ & $90.5_{0.4}$ & 88.6 \\ \cline{2-9}
& Relative Gain & $\mathbf{0.313_1}$ & $\underline{\mathbf{0.317_{1,2}}}$ & $\mathbf{0.297_1}$ & $\mathbf{0.295_1}$ & $\mathbf{0.280_1}$ & 0.258 & \textbf{0.293} \\
\hline \hline
\multirow{7}{*}{\textbf{WordNet}} & Utility F1 & $84.7_{0.5}$ & $83.1_{1.6}$ & $83.4_{1.0}$ & $83.7_{2.2}$ & $85.0_{0.8}$ & $86.8_{1.7}$ & 84.4 \\
& CS & 0.876 & 0.865 & 0.863 & 0.875 & 0.781 & 0.962 & 0.870 \\
& Perplexity & 159 & 162 & 162 & 159 & 177 & 96 & 152 \\ \cline{2-9} 
& PI Masking & 17.71 & 18.04 & 17.70 & 18.05 & 18.57 & 28.89 & 19.83 \\
& Static F1 & $74.9_{2.0}$ & $72.5_{2.5}$ & $73.7_{2.8}$ & $74.9_{2.0}$ & $71.7_{2.1}$ & $88.4_{0.8}$ & 76.0 \\
& Adaptive F1 & $88.8_{0.9}$ & $89.4_{0.8}$ & $88.5_{1.2}$ & $88.5_{0.6}$ & $89.0_{1.0}$ & $92.3_{1.0}$ & 89.4 \\\cline{2-9} 
& Relative Gain & $\mathbf{0.290_2}$ & 0.280 & 0.281 & 0.283 & 0.248 & 0.249 & 0.272 \\
\hline
\multirow{7}{*}{\textbf{Average}} & \textbf{Utility F1} & 84.4 & 83.4 & 82.9 & 83.5 & 83.4 & 86.0 & -- \\
 & \textbf{CS} & 0.881 & 0.869 & 0.869 & 0.878 & 0.797 & 0.959 & -- \\
 & \textbf{Perplexity} & 164 & 166 & 167 & 164 & 180 & 103 & -- \\ \cline{2-9} 
 & \textbf{PI Masking} & 19.92 & 20.15 & 19.84 & 20.11 & 20.46 & 27.94 & -- \\
 & \textbf{Static F1} & 72.8 & 70.3 & 71.2 & 73.0 & 70.0 & 87.0 & --\\
 & \textbf{Adaptive F1} & 88.8 & 88.6 & 88.9 & 88.8 & 88.9 & 91.4 & -- \\ \cline{2-9} 
 & \textbf{Relative Gain} & \textbf{0.289} & 0.287 & 0.281 & 0.283 & 0.247 & 0.254 & -- \\
\hline
\end{tabular}
}
\caption{Results for the Yelp dataset at $\varepsilon=935$. Each combination of decomposition and distribution strategy yields scores for Utility F1 ($\uparrow$), Cosine Similarity (\textit{CS}, $\uparrow$), Perplexity ($\downarrow$), PI Masking ($\downarrow$), Static F1 ($\downarrow$), Adaptive F1 ($\downarrow$), and Relative Gain ($\uparrow$). Values derived from an average of three runs are displayed as mean$_{\text{std}}$. F1 values are in \%. The rightmost column and bottom rows calculate the average results over decomposition and distribution methods, respectively. Best average Relative Gain values are \textbf{bolded}. For individual Relative Gain values, only the best are bolded: $\mathbf{x}_1$ for best per decomposition, $\mathbf{x}_2$ for best per budget distribution, and $\mathbf{x}_{1,2}$ if both. The global best Relative Gain (across all combinations) is \underline{underlined}.}
\label{tab:results_trustpilot_935}
\end{table*}

\section{Example Texts.}
Table \ref{tab:examples1} shows corresponding text examples from a single original text from the Trustpilot dataset, privatized with all combinations of decomposition and distribution methods, at $\varepsilon = 52$.

\begin{table*}[ht!]
    \centering
    \scriptsize
    \resizebox{\linewidth}{!}{
\begin{tabular}{c|c|p{1.1\textwidth}}
\multirow{3}{*}{\textbf{Decomp.}} & \multirow{3}{*}{\textbf{Dist.}} & \textit{Great website, products and customer service!: They we great. Delivery was 7-7 but the text me a smaller window the night before so. Didn't have to stay  in all day which was great and they called when they were ten mins away. One item was cracked but when I called up I didn't have to sit on hold or go through a complicated automated system, I got straight through to someone who arranged to send another item out ASAP. Really pleased} \\ \hline
\multirow{23}{*}{PMI} & Baseline & anthemic help create, and products or services!: they we nice. may still was career average - but the attorneys me a car club unable to move the penalty minutes before so. didn't have to gotten in all contact form which was fields and they attempt was made when they were research project canonical. great outdoors was inclusion criteria but when i list up i didn't have to organizers on seamless or i'll through a downloaded windows registry, i cursor close eye through to opened fire who nerdwallet to tenofovir sure everything out field. fun to make \\
 & Attention & funeral home chapel million in revenue, and products and services!: they we sure that everyone. fields like was within 30 - but the conversation going me a penalty exclusive license the round pick before so. didn't have to halftime in all jel classification which was late 1800 and they annex when they were lieutenants identifiers. clothing company was would be perfect but when i deforestation up i didn't have to hurl on things going or google account through a runtime effortless, i really liked second goal through to minaj who would fit to stipulated big bang theory out available on itunes. harshly \\
 & Gradients & relegated family style, and higher revenue!: they we snow. started writing was new poll - but the tracking number me a wagering requirements primary research the looked amazing before so. didn't have to feasibility study in all scheduled for release which was settings page and they consisting when they were pastures time to put. html file was awarded but when i cram up i didn't have to conversation going on regions or choose us through a cute extensive range, i bought jangle through to sure everyone who encrypted to shop childrens work sure everyone out wondering if anyone. really stood \\
 & IC & would make a great wipo case, and available online!: they we footprint. questionnaire was 2 2 - but the fungus me a 4gb ever worked the vibrate before so. didn't have to frequently asked questions in all horsepower which was without the need and they also more likely when they were immortalized shop childrens work. heaviness was parisian but when i hectares up i didn't have to open mind on uphold or strong sense through a metrology elegantly, i hand away easily without through to posts rr who could use to smashwords vsphere out mechanical ventilation. really happy \\
 & \textsc{KeyBERT} & momentum going get organized, and screenshots!: they we babip. online survey was nov - but the easily me a rapidly pokemon the auction before so. didn't have to eventually become in all quilt which was minimalism and they mansion when they were tibetan min action adventure. default was hip hop but when i paved the way up i didn't have to start the day on anxiety or fun playing through a other streak, i pm powered by vbulletin sarah jessica parker through to fits well who seater to whole or in part offer a full out leach. pleaded guilty to one \\
 & \textsc{YAKE} & included studies mail buy, and client service!: they we editors. craig1916 was 7 7 - but the tolstoy me a long and short samsung galaxy the side before so. didn't have to clock ticking in all dresses which was presided and they localhost when they were terre apparent attempt. purpose flour 1 was features built but when i decided to use up i didn't have to sit on hold or go through a complicated automated system, i got straight through to someone who arranged to send another 2019 03 out root access. username and password \\ \hline
\multirow{23}{*}{LLR} & Baseline & ungodly website, anti theft system google announced!: they we p m cst. ll was consumers may - 120hz but the apes me a helping the anyplace before so. didn't have to yet received in all awarded which was please and they peacekeeping operations when they were franchise history best outdoor. genetic counseling was meet the requirements but when i imprisoned up i didn't have to sitting on term life insurance or account with us through a complete stop story was originally, i second attempt pollen through to sore muscles who pushbutton to give rise served as the director out domain name. maceration \\
 & Attention & software program zip file, data cannot insurer!: they we conservator. finances was dating app - stocking but the akc me a full frame metacritic the feel free before so. didn't have to take care in all took us which was member of the board and they let us when they were second period keyless entry. maps was co v but when i please feel free up i didn't have to weight room on best practices or grow through a fermentation ecommerce platform, i crying electric shock through to girl who knew i wanted to key fob food stalls out dedicated to helping. rear cross traffic \\
 & Gradients & please payback period, across the web calculate!: they we winnings. yet released was 1xbet - push button but the termed me a isdn automatic the plate appearances before so. didn't have to forefoot in all motherboard which was final whistle and they justice when they were sarajevo nest. outbreak of war was parked but when i burned up i didn't have to straighten on video formats or reels through a recently received crypto currency, i panoramic sunroof swee through to remi maintenance fee reminder who wanted to see to might make drumsticks out cannot guarantee. made me realize \\
 & IC & screen canva, last moment line casino!: they we it'd. win one was annual production - little did i know but the readmissions me a rooms are equipped several attempts the ajga before so. didn't have to left to play in all morning which was mocking and they fascinating when they were professor nair. elantra was two day but when i retarded up i didn't have to sunday april on engineering manager or feel free through a struggling amount of water, i marries lame through to water flow who punished to actual damages another vehicle out six figures. year in a row \\
 & \textsc{KeyBERT} & tailgate please feel free, last summer unsuccessful attempt!: they we mobs. yet received was seemed like - yet reached but the remote keyless entry me a situated eliminate the 970 before so. didn't have to international airports in all discipleship which was fun and they attempts when they were received run of the mill. odoo was appoint but when i made it possible up i didn't have to yell on take or get to go through a compares type of treatment, i previous clients 100 0 through to something who better person to please feel free gmo free out pills. purposed \\
 & \textsc{YAKE} & if time and cost, strange at first data protection officer!: they we plugin. distributed was 7 - 7 but the independent contractor me a opting first try the friday night before so. didn't have to stay in all week since which was fun and they mm thick when they were g dl author david. discounted was clunky but when i lean startup up i didn't have to sit on hold or go through a complicated automated system, i got straight through to someone who arranged to send cares act out disappearance. really pleased \\ \hline
\multirow{20}{*}{t-score} & Baseline & term contract, 13mp!: they we bright future. pdt was gulps - but the median follow me a continue to grow materials the last days before so. didn't have to sects in all success which was always a good and they common cause when they were curse words get inspiration. construction debris was principalities but when i incubator up i didn't have to lay their eggs on gently or heart through a nests commercial drivers, i two girls transfer rate through to nugget who graze to large mixing overlap out academically. hard to get \\
 & Attention & brief conversation, free sites!: they we caffe. forms including was 67 - but the selections me a topics including feet of water the world before so. didn't have to christmas in all greasy which was something and they uk s leading when they were reservations with confidence days to go. certificates of deposit was episode but when i recently asked up i didn't have to laugh on completed within or going to take through a hard to come time mixing, i feet high material including through to loved one who miles southwest to heavenly host prior written consent out happy life. bringing people together \\
 & Gradients & pooja, bombay!: they we gamescom. reforms was episode - but the christmas special me a hotels honourable the closing before so. didn't have to resins in all eighth inning which was crates and they area known when they were hard to take players will also. sturdiness was barry bonds but when i elapsed up i didn't have to something on find many or spaced through a full potential early to say, i levees villain through to amicable who 1926 to excavation future growth out dad. jla \\
 & IC & sprout, mobility issues!: they we yearbook. christmas was kilometers west - but the page numbers me a hamilton beach centre the rishikesh before so. didn't have to new episode in all entire life which was many products and they could live when they were aired fastest growing companies. every name was oozed but when i quae up i didn't have to moment on mpeg or think people through a hundred feet full potential, i teammates fairway through to sources including who plastered to inquiring items including out arthritic. therefore essential \\
 & \textsc{KeyBERT} & balanced approach, characters including!: they we work closely. reputation was supporter - but the items including me a much more fun asbestos removal the sinai before so. didn't have to easier to see in all memories which was materials like and they uttered when they were please mention this item 11 00 11. valid reason was guru but when i household name up i didn't have to better to go on genesis 1 or buy through a intricate copyrights, i selection of games upstarts through to child who recipe calls to speak to us competencies out hard cock. new team \\
 & \textsc{YAKE} & blog article, cmx!: they we yards and a touchdown. products including was 7 7 - but the text me a traverse oiling the cherry trees before so. didn't have to area near in all pagoda which was materials including and they blues when they were dispersed claude monet. prostitutes was terminator but when i application number us up i didn't have to sit on hold or go through a complicated automated system, i got straight through to someone who arranged to send another ebay out us to share. aspects of the game \\ \hline
\multirow{18}{*}{POS} & Baseline & great website products, and customer service!: they we 2019 20. was kelvin - but the link back me a fir the wick before so. didn't have to there's in all baking which was fragile and they tossed when they were network traffic appropriate action. medicine cabinet was free trade agreement but when i instrumental up i didn't have to sprung on crossed or exists through a complicated automated system, i angrily through to candidate who first paragraph spill out listeners. special issue first product \\
 & Attention & great website products, and washing!: they we new file. was 7 4 - but the valid license me a planning to put the barking before so. didn't have to skin feels in all entire month which was cold front and they first full season when they were good fortune stairs. jennifer aniston was humidity but when i opened up i didn't have to sitting on convey or one god through a complicated automated system, i insults through to reasonable steps who designed to attract funny story out educators. quran indian institute \\
 & Gradients & great website products, and leadership team!: they we slinging. was reigning - but the dug me a moment the supervise before so. didn't have to diocese in all basic needs which was pumped and they motorists when they were insects away. health reasons was carved but when i colonial up i didn't have to multiple choice questions on blood sample or raged through a complicated automated system, i harried through to reasonable efforts who entire list drab out grew. simple easy secretary \\
 & IC & great website products, and unison!: they we documents related. was online sources - but the christmas morning me a expansion the old one before so. didn't have to relevant references in all nfl draft which was cannot rely and they global trade when they were trampling treadmill. picasso was lot bigger but when i buddhism up i didn't have to body composition on also receive or put through a complicated automated system, i decided to drive through to law who tiptoe fluorine out tallied. slowed loving god \\
 & \textsc{KeyBERT} & great website products, and protests!: they we marshal. was oppo - but the legal obligation me a failed to act the sex offenders before so. didn't have to drying time in all bottle which was rich and they ate when they were grasses optioned. two glasses was offside but when i attacked up i didn't have to greet on unconscious mind or stumbles through a complicated automated system, i ask to speak through to one song who gross margin similar product out cooperation. never use happy \\
 & \textsc{YAKE} & great website products, and customer service!: they we international federation. was 7 7 - but the text me a search keyword the night before so. didn't have to working directly in all massacred which was slowdown and they jammed when they were healthy gums away. someone was threw but when i last second up i didn't have to sit on hold or go through a complicated automated system, i got straight through to someone who arranged to send live and work out would taste. really commercialize \\ \hline
\multirow{18}{*}{WordNet} & Baseline & loopholes everest, prizes and subscribe service!: they we lacking. bingo was travelled - sleeps but the odometer me a fiscal years number the evening before so. didn't have to spins in all 10 which was snooze and they there'll when they were jazz lunch montgomery. trial shipment was dewy but when i focuses up i didn't have to poker games on shock waves or fiscal years through a objected alarming chassis, i finished antennae through to snooze who lacking to remind they'll securities and exchange commission out phys. very fiscal years \\
 & Attention & bomber take advantage, 1xbet and inefficiencies insured!: they we smoother. raimi was 24 - 256 but the paylines me a perpetrate lacking the fiscal years before so. didn't have to keep in all months which was perfect and they undermines when they were rockets wonderful fibers. regiment solution was poked but when i wanna up i didn't have to expressway on gambling or copperas through a home buyer mission impossible infantry, i hung roof through to them who next door to donate accusation financial year out melinda. prove truth \\
 & Gradients & yosemite who'll, raimi and brewery lenders!: they we fantastic. locomotives was ean - trials but the townland me a there italian renaissance the nightclub before so. didn't have to stay in all preview which was wysiwyg and they tyne when they were decreased pokies unmoved. ecole electrolyte was bet but when i eerie up i didn't have to sit on announced or kings through a sharper iphone bonuses, i gets got through to truth who trial to confirm latest recipe out gaming. snooze shockwaves \\
 & IC & manicured warship, products and 4in wi!: they we achieved. dont was 16 - 11 but the covariates me a sensitive fy the uncle before so. didn't have to reunite in all rows which was sparked and they fallacy when they were constantinople min loaded. filling reels was eraser but when i shattered up i didn't have to sit on slots or ginseng through a challenges proofing frequently, i lived outboard through to difficulty who aimed to send feige slots out gaming. revengeance msci \\
 & \textsc{KeyBERT} & cope federal trade commission, games and breaks birth control!: they we humanistic. million was 00 - teaches but the experiments me a rocky usdfor the night before so. didn't have to tread in all regiment which was good time and they holy week when they were 170 feel pleasant. lacking calendar year was we'll but when i oppose up i didn't have to sit on wait or send through a possible let's secures, i fades water tanks through to sports car who wronged to respond impermanence construed out sto. drug addict misdiagnosed \\
 & \textsc{YAKE} & 1xbet website, items and aftermarket service!: they we lot. mailbox was 7 - 7 but the manuela me a sporty window the weeknight before so. didn't have to traverse city in all norte which was corsets and they adjudicating when they were five mins away. one thermometer was cracked but when i yelling up i didn't have to sit on hold or go through a complicated automated system, i got straight through to someone who arranged to send another indictment out they'll. really pleased
\end{tabular}
}
\caption{Obfuscated output examples from the Trustpilot dataset, at a document-level $\varepsilon = 52$.}
\label{tab:examples1}
\end{table*}

\end{document}